\documentclass[letterpaper]{article} 

\usepackage[preprint]{aaai2027}
\usepackage[hyphens]{url}  
\usepackage{graphicx} 
\usepackage{natbib}  
\usepackage{caption} 
\usepackage{algorithm}
\usepackage{algorithmic}

\usepackage{newfloat}
\usepackage{listings}
\DeclareCaptionStyle{ruled}{labelfont=normalfont,labelsep=colon,strut=off} 
\floatstyle{ruled}
\newfloat{listing}{tb}{lst}{}
\floatname{listing}{Listing}

\usepackage{booktabs}

\usepackage{amsmath}
\usepackage{amssymb}

\usepackage{array}

\title{Learning Context-Aware Motion Priors for Humanoid Control}

\author{
    Yunyang Mo\textsuperscript{\rm 1}\equalcontrib,
    Yi Gu\textsuperscript{\rm 1}\equalcontrib,
    Yangchen Zhou\textsuperscript{\rm 1}\equalcontrib,
    Hanyang Cao\textsuperscript{\rm 1},
    Renjing Xu\textsuperscript{\rm 1}\corresponding
}

\affiliations{
    \textsuperscript{\rm 1}The Hong Kong University of Science and Technology (Guangzhou),
    Guangzhou, China\\
    renjingxu@hkust-gz.edu.cn
}

\begin{document}
\maketitle
\begin{abstract}
Motion priors provide powerful guidance for learning naturalistic humanoid behaviors. However, existing methods typically learn a general, task-agnostic prior from the entire reference dataset and apply it uniformly throughout policy training. As a result, the prior cannot distinguish which reference motions are relevant to the current task context, potentially providing irrelevant or conflicting guidance. We present Context-Aware Motion Priors (CMP), a framework that adapts a general motion prior to the current task context without manual skill labels, dataset partitioning, or a separate skill discovery stage. Specifically, CMP learns context--motion compatibility using high-advantage policy rollouts, while a demonstration-based objective keeps the learned relevance grounded in the reference distribution. The resulting relevance scores reweight reference supervision for training a lightweight context-conditioned adapter. To evaluate the effectiveness and generality of CMP, we instantiate it with both Adversarial Motion Priors and Score-Matching Motion Priors. Across five humanoid control tasks, CMP consistently improves task performance and sample efficiency, learns meaningful context--motion alignment, and remains robust to imbalanced reference distributions. These results show that adapting motion priors to task contexts provides more relevant guidance for humanoid policy learning.
\end{abstract}

\section{Introduction}
\label{sec:intro}
Motion priors have become a central component of humanoid control based on reinforcement learning. By encouraging the policy to generate motions that resemble human demonstrations, methods such as Adversarial Motion Priors (AMP)~\cite{peng2021amp} enable simulated characters and humanoid robots to acquire natural, physically plausible behaviors while optimizing downstream task objectives. As humanoid controllers become more versatile, however, motion priors are increasingly trained on heterogeneous collections spanning different speeds, directions, styles, transitions, and behaviors. Although such diversity broadens behavioral coverage, existing priors primarily model whether a motion is plausible in general, rather than whether it is relevant to the current task context defined by the goal, command, future trajectory, or environment state.

\begin{figure}[t]
    \centering
    \includegraphics[width=\linewidth]{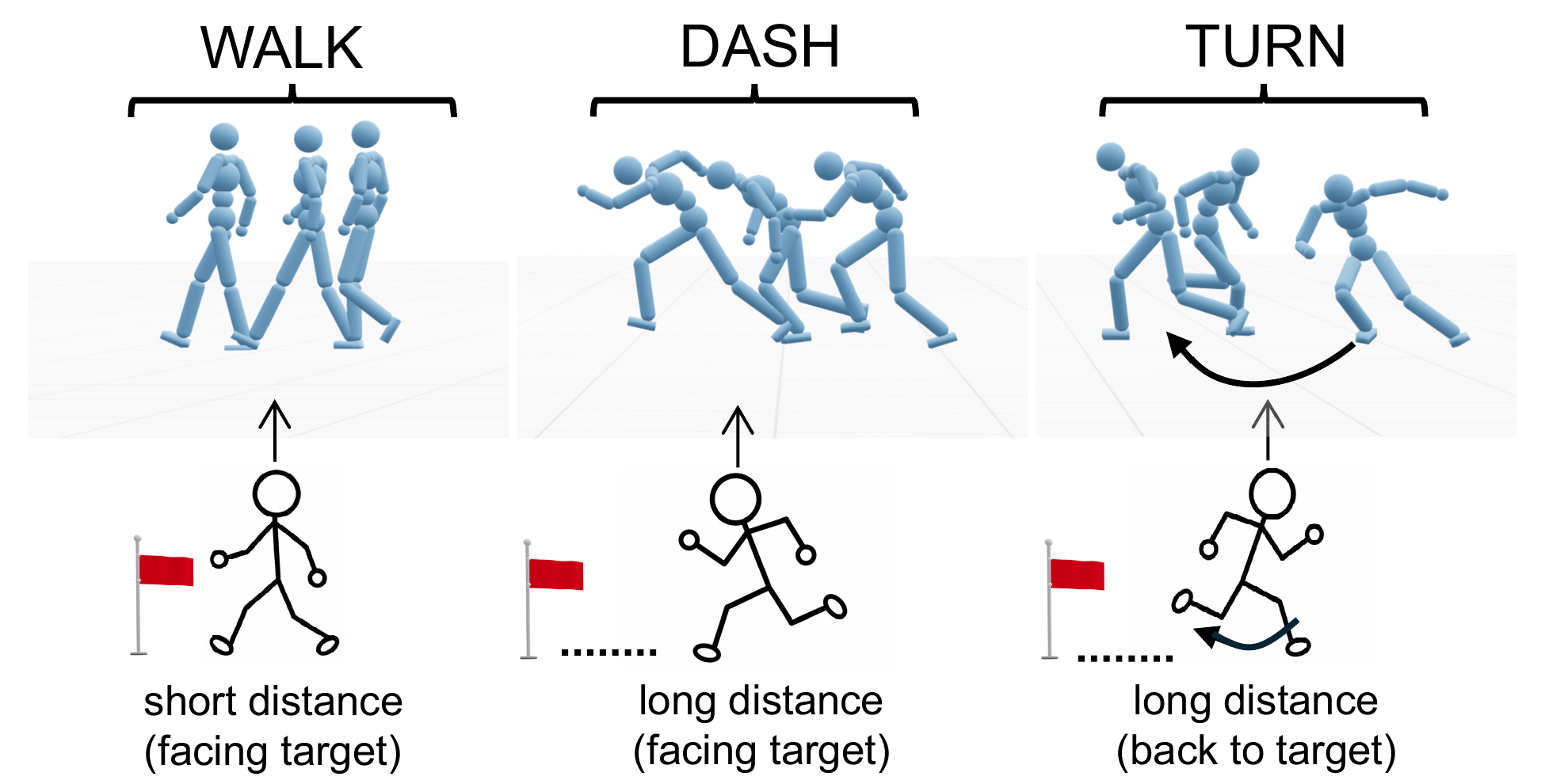}
    \caption{\textbf{Reference motions for reaching a target location.}
Different reference motions, such as walking, dashing, and turning, may
provide more suitable guidance depending on the humanoid's distance and
orientation relative to the target.}
    \label{fig:teaser}
\end{figure}

Motion plausibility and task relevance are not equivalent. As illustrated in Fig.~\ref{fig:teaser}, reaching a nearby target may favor walking, whereas reaching a distant target may require faster locomotion; if the target is behind the humanoid, turning may be necessary before moving forward. Similar dependencies arise from velocity commands, future trajectories, incoming objects, and interaction states. Conventional motion priors nevertheless use the same reference distribution across task contexts. Consequently, motions that are physically plausible but poorly matched to the current task may provide conflicting supervision, slow policy optimization, and bias the learned behavior toward frequent motion modes, particularly when the reference dataset is heterogeneous or imbalanced.

Prior work addresses motion heterogeneity primarily by learning latent skills or behavior modes and allowing a downstream controller to select among them~\citep{ase,case,calm}. While effective, these approaches introduce an intermediate skill space that is typically learned before downstream control and remains fixed thereafter. We instead ask: \textbf{Can a general motion prior adapt online to the current task context directly in the original reference space, without skill labels, dataset partitioning, or a separate skill-discovery stage?
}

We introduce \emph{Context-Aware Motion Priors} (CMP), a modular framework that learns the compatibility between task contexts and motion sequences from policy experience. CMP uses high-advantage policy rollouts to identify motions that are useful in each task context. Because policy rollouts collected during online exploration may not fully exhibit the behaviors represented in the reference dataset, we introduce a demonstration-positive objective to anchor the learned context representation to the reference motion distribution.

The compatibility scores define soft relevance weights over reference motions, continuously reshaping the effective reference distribution for each task context. CMP uses the weights to train a lightweight context-conditioned residual adapter while leaving the original training procedure of the base motion prior unchanged. This formulation separates general motion plausibility, represented by the base prior, from context-dependent task relevance, captured by the adapter. CMP can be integrated with different motion-prior formulations. We instantiate CMP with Adversarial Motion Priors~\cite{peng2021amp} as our primary setting and extend it to Score-Matching Motion Priors~\cite{mu2025smp}.

We evaluate CMP across five simulated humanoid control tasks in which the task context varies with spatial goals, movement commands, future trajectories, or object states. Across all five tasks and both motion-prior formulations, CMP delivers consistent gains in task performance and sample efficiency, demonstrating the broad applicability of context-aware prior adaptation. Analyses of the learned relevance weights and retrieved motions reveal meaningful context--motion alignment, while ablations verify the contributions of advantage-based relevance learning and demonstration support. CMP also remains robust when the reference distribution is made severely imbalanced, indicating that context-aware adaptation reduces sensitivity to dominant reference frequencies.

Our contributions are summarized as follows:
\begin{itemize}
    \item We identify a fundamental mismatch in existing motion priors between context-independent motion plausibility and context-dependent task relevance, and formulate the use of heterogeneous reference data as a context-conditioned relevance problem.
    \item We introduce \emph{Context-Aware Motion Priors}, which learn context--motion compatibility from high-advantage policy experience together with demonstration-supported regularization, and use the resulting relevance weights to train a lightweight context-conditioned adapter without motion labels, dataset partitioning, or hard reference selection.
    \item We validate CMP on five humanoid control tasks and two distinct motion-prior formulations, demonstrating consistently improved task performance and sample efficiency, interpretable context–motion alignment, and robustness to severe reference-distribution imbalance.
\end{itemize}

\section{Related Work}

\subsection{Motion Priors for Humanoid Control}

Motion priors guide task policies toward human-like behaviors without
explicitly tracking a prescribed motion trajectory. AMP learns an
adversarial prior by distinguishing policy-generated motions from
reference data~\citep{peng2021amp}, while SMP constructs reusable,
task-agnostic priors from pretrained diffusion
models~\citep{mu2025smp}. Despite their different formulations, both
provide guidance from the reference distribution as a whole, without
modeling how each motion's relevance changes with the current task context.

Another line of work represents heterogeneous motion data through reusable
latent skills. ASE learns skill embeddings from unstructured motion
clips~\citep{ase}, CALM learns semantic latent controls for directable
behaviors~\citep{calm}, and C$\cdot$ASE partitions heterogeneous motions
into homogeneous subsets to learn conditional behavior
distributions~\citep{case}. \citet{luo2024universal} distill an
imitator trained on large unstructured datasets into a universal latent
representation for downstream control. These methods formulate motion
diversity as a latent skill representation and selection problem. CMP
instead retains the original reference space and learns context-dependent
relevance during downstream learning, adapting each reference motion's
contribution within the motion-prior objective.
\subsection{Adaptive Reference Selection}

Prior work adapts reference sampling to imitation difficulty by prioritizing underlearned motions or segments based on tracking performance, failure frequency, or transition difficulty~\citep{klipfel2023learning,yu2025skillmimic,pan2025agility,liao2025beyondmimic,chen2025gmt}.
Motion matching and its learned variants retrieve or approximate the
retrieval of motions based on pose, trajectory, or environment features
~\citep{clavet2016motion,bergamin2019drecon,holden2020learned,
ponton2025environment}.
Although effective, these methods rely on imitation difficulty or hand-designed matching costs rather than task relevance learned through downstream reinforcement learning.

CMP learns task-conditioned relevance from advantage signals collected
during online policy learning. It applies the resulting soft weights
directly within the motion-prior objective, adapting the effective
reference distribution to the current task context without hard retrieval
or prioritizing references solely according to tracking difficulty.
\subsection{Contrastive Representation Learning for Control}

Contrastive learning organizes representations by bringing compatible pairs
closer while separating mismatched pairs
~\citep{oord2018representation,radford2021learning}. In reinforcement
learning, contrastive objectives have been used for visual representation
learning~\citep{laskin2020curl}, future-state reachability, and
goal-conditioned value functions
~\citep{eysenbach2021c,eysenbach2022contrastive}, and have been further
adapted to offline robotic goal reaching
~\citep{zheng2024stabilizing}.

CMP applies contrastive learning to a different compatibility relation.
Rather than estimating whether a state can reach a goal, CMP measures whether
a reference motion is relevant to the current task context. The learned
context--motion scores are then used to reweight reference supervision within
the motion-prior objective.

\section{Method}
\label{sec:method}

\begin{figure}[t]
    \centering
    \includegraphics[width=1.0\linewidth]{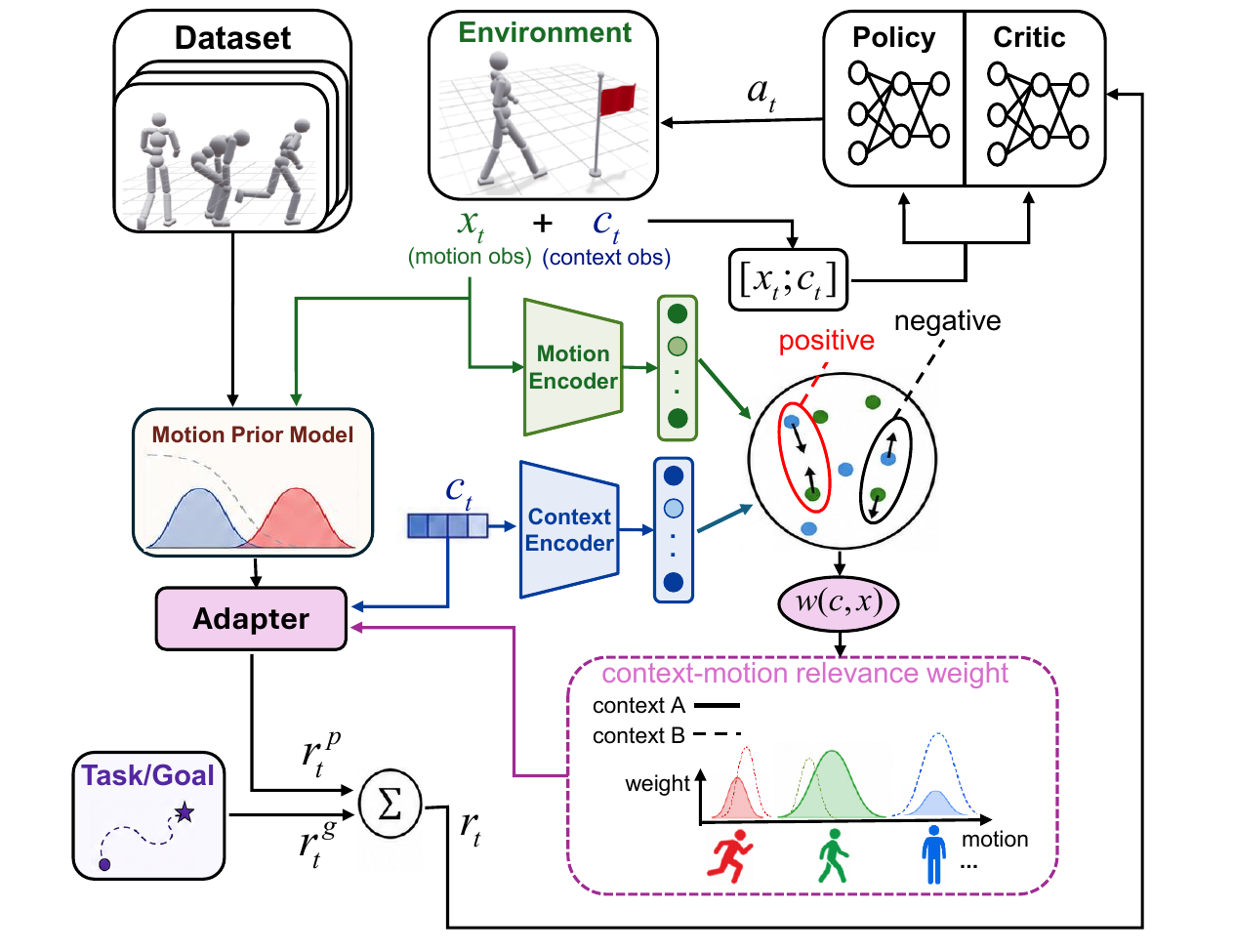}
    \caption{\textbf{Overview of CMP.}
CMP learns context--motion relevance and uses it to train a
relevance-weighted context adapter over the base motion prior.}
    \label{fig:pipeline}
\end{figure}

Context-Aware Motion Priors (CMP) is a modular framework that augments
task-agnostic motion priors with context-dependent reference motion
relevance, enabling adaptive use of motion references for different tasks.
As illustrated in Fig.~\ref{fig:pipeline}, CMP learns a contrastive relevance function
between task contexts and reference motions. The resulting scores induce
an adaptive reweighting of the reference distribution and guide the
adaptation of the motion prior to the current task context.
We first introduce the reweighting interpretation of CMP in
Section~\ref{subsec:cmp_reweighting}, then describe the contrastive
relevance learning procedure in Section~\ref{subsec:contrastive_relevance}.
Finally, we instantiate CMP with AMP and extend it to SMP in
Section~\ref{subsec:integration_motion_priors}.

\subsection{A Reweighting View of Context-Aware Motion Priors}
\label{subsec:cmp_reweighting}

Let $\mathcal{D}_E=\{x_n\}_{n=1}^{N}$ denote a reference motion
dataset, where $x\in\mathcal{X}$ is a motion observation sequence, and
let $p_E(x)$ be its empirical distribution. We consider the
reference-side term of a task-agnostic motion-prior objective:
\[
\mathcal{L}_{\mathrm{ref}}(\theta)
=
\mathbb{E}_{x\sim p_E}
\left[
\ell_{\theta}^{\mathrm{ref}}(x)
\right].
\]
 Under this objective, the contribution of each reference motion
is determined by $p_E$ and is independent of the current task context.

CMP instead learns a relevance function
\[
R_{\phi}:\mathcal{C}\times\mathcal{X}\rightarrow\mathbb{R},
\]
where $c\in\mathcal{C}$ denotes the current task context. The relevance
score induces a normalized importance weight:
\[
w_{\phi}(c,x)
=
\frac{
\exp\!\left(\alpha R_{\phi}(c,x)\right)
}{
\mathbb{E}_{x'\sim p_E}
\left[
\exp\!\left(\alpha R_{\phi}(c,x')\right)
\right]
},
\]
where $\alpha>0$ controls the sharpness of the reweighting. Under this
population-level definition,
$\mathbb{E}_{x\sim p_E}[w_{\phi}(c,x)]=1$, yielding the effective
context-conditioned reference distribution
\[
q_{\phi}(x\mid c)
=
w_{\phi}(c,x)p_E(x).
\]
This distribution assigns greater importance to reference motions that
are more compatible with the current context.

In practice, we use an efficient minibatch approximation. Given paired
contexts and reference motions
$\mathcal{B}=\{(c_i,x_i)\}_{i=1}^{B}$, we compute
\[
\bar w_\phi(c_i,x_i)
=
\operatorname{clip}_{[w_{\min},w_{\max}]}\!\left(
\frac{B\exp(\alpha R_\phi(c_i,x_i))}
{\sum_{j=1}^{B}\exp(\alpha R_\phi(c_i,x_j))}
\right).
\]
For each context, normalization gives the weights unit mean over the
reference minibatch; subsequent clipping bounds them for stable training.

CMP trains a context-conditioned adapter with parameters $\psi$ using
detached relevance weights:
\[
\mathcal{L}_{\mathrm{adapt}}(\psi\mid\theta,\phi)
=
\mathbb{E}_{\substack{c\sim d_{\pi}\\x\sim p_E}}
\left[
\operatorname{sg}\!\left[\bar{w}_{\phi}(c,x)\right]
\ell_{\psi}^{\mathrm{adapt}}(c,x;\theta)
\right],
\]
where $d_{\pi}$ denotes the context distribution induced by the current
policy. The reference motions $x \sim p_E$ can be reused from the same
minibatch sampled for $\mathcal{L}_{\mathrm{ref}}$. Gradients from this
objective update only the adapter parameters $\psi$; they do not propagate
into the relevance model $\phi$ or the base prior $\theta$. The base prior may nevertheless continue to be optimized by its original training objective.

This practical objective approximates training on reference motions drawn
from $q_{\phi}(x\mid c)$. CMP thus adapts the effective reference
distribution without modifying the reference dataset or the underlying
prior. The full adapter objective and the exact population-level
equivalence are detailed in Appendix~A.

\subsection{Contrastive Relevance Learning}
\label{subsec:contrastive_relevance}

The reweighting view in Section~\ref{subsec:cmp_reweighting} assumes a
relevance function $R_{\phi}(c,x)$ that measures the compatibility between
a task context $c$ and a motion sequence $x$. We next describe how CMP learns this
function from online policy rollouts and reference motions, without
explicit context--motion compatibility annotations.

Let $f_{\phi_c}$ and $g_{\phi_x}$ denote the context and motion encoders,
respectively, where $\phi=(\phi_c,\phi_x)$. We define
\[
R_{\phi}(c,x)
=
\frac{
f_{\phi_c}(c)^{\top}g_{\phi_x}(x)
}{
\|f_{\phi_c}(c)\|_2\,
\|g_{\phi_x}(x)\|_2
},
\]
and
\[
s_{\phi}(c,x)
=
\exp\!\left(R_{\phi}(c,x)/\tau\right),
\]
where $\tau>0$ is the contrastive temperature.

\paragraph{Online task-relevance branch.}
We first extract task-relevance signals from online policy rollouts. Let
\[
\mathcal{B}_{\mathrm{on}}
=
\{(c_i,x_i,\hat{A}_i)\}_{i=1}^{B}
\]
denote a mini-batch of rollout samples, where $\hat{A}_i$ is the standardized
Generalized Advantage Estimate (GAE). We select high-advantage samples as
positive context--motion anchors:
\[
\mathcal{I}_{\mathrm{on}}
=
\{i\mid\hat{A}_i>0\}.
\]
For each $i\in\mathcal{I}_{\mathrm{on}}$, the paired motion $x_i$ is treated
as positive for context $c_i$, while other motions in the batch serve as
contrastive candidates. The online loss is
\[
\begin{aligned}
\mathcal{L}_{\mathrm{on}}(\phi)
=
-\frac{1}{|\mathcal{I}_{\mathrm{on}}|}
\sum_{i\in\mathcal{I}_{\mathrm{on}}}
\rho_i
\log
\frac{s_{\phi}(c_i,x_i)}
{\sum_{j=1}^{B}s_{\phi}(c_i,x_j)} .
\end{aligned}
\]
Here, $\rho_i=\sigma(\hat{A}_i/\beta_{\mathrm{adv}})$ weights each positive
pair according to its advantage, and $\beta_{\mathrm{adv}}$ controls the
scaling. If too few samples satisfy $\hat{A}_i>0$, we instead select the
highest-advantage samples in the batch.

\paragraph{Demonstration-positive branch.}
The online branch identifies rollout motions that are effective for the
current task, but online exploration may not fully exhibit the behaviors
represented in the reference dataset.
We therefore introduce a demonstration-positive branch to anchor relevance
learning to supported motions. For the same context batch
$\{c_i\}_{i=1}^{B}$, we sample reference motions
$\mathcal{B}_E$ and denote the online rollout motions by
$\mathcal{X}_{\mathrm{on}}=\{x_i\}_{i=1}^{B}$. The resulting loss is
\[
\begin{aligned}
\mathcal{L}_{\mathrm{demo}}(\phi)
=
-\frac{1}{B}
\sum_{i=1}^{B}
\log
\frac{
\sum_{x_E\in\mathcal{B}_E}s_{\phi}(c_i,x_E)
}{
\sum_{x\in\mathcal{B}_E\cup\mathcal{X}_{\mathrm{on}}}
s_{\phi}(c_i,x)
}.
\end{aligned}
\]
Although positive in $\mathcal{L}_{\mathrm{on}}$, rollout motions are
negative here: the online branch captures task relevance, whereas the
demonstration branch anchors the representation to reference-supported
motions without imposing task-specific preferences among them. The final relevance objective is
\[
\mathcal{L}_{\mathrm{rel}}(\phi)
=
\mathcal{L}_{\mathrm{on}}(\phi)
+
\lambda_{\mathrm{demo}}
\mathcal{L}_{\mathrm{demo}}(\phi),
\]
where $\lambda_{\mathrm{demo}}$ controls the strength of the demonstration-positive branch.

\subsection{Integration with Motion Priors}
\label{subsec:integration_motion_priors}

We first instantiate the adapter loss
$\ell_{\psi}^{\mathrm{adapt}}(c,x;\theta)$ defined in
Section~\ref{subsec:cmp_reweighting} for AMP, which serves as our primary
setting, and then extend the same formulation to SMP. In both cases,
detached relevance weights $\bar{w}_{\phi}(c,x)$ are applied only to the
reference-side adapter objective, and gradients from this objective do not
propagate into the base motion prior.

\paragraph{Primary Instantiation with AMP.}
Let $l_{\theta}(x)\in\mathbb{R}$ denote the scalar logit produced by the
base AMP discriminator. CMP defines the context-conditioned logit as
\[
l_{\theta,\psi}(c,x)
=
\operatorname{sg}\!\left[l_{\theta}(x)\right]
+
\lambda_{\mathrm{res}}\Delta l_{\psi}(c,x),
\]
where $\lambda_{\mathrm{res}}$ controls the adapter strength. This
instantiates the generic adapter formulation in
Section~\ref{subsec:cmp_reweighting} and
Appendix~A with
$z_{\theta}=l_{\theta}$ and
$\Delta_{\psi}=\Delta l_{\psi}$. The base discriminator continues to be
trained through the original AMP objective, which is reviewed in
Appendix~B.

For reference motions, the prior-specific loss is a relevance-weighted
binary cross-entropy loss with target label one:
\[
\mathcal{L}_{\mathrm{AMP}}^{+}(\psi)
=
\mathbb{E}_{\substack{c\sim d_{\pi}\\x_E\sim p_E}}
\left[
\bar{w}_{\phi}(c,x_E)
\operatorname{BCE}
\left(l_{\theta,\psi}(c,x_E),1\right)
\right].
\]

Policy-generated motions retain an unweighted negative term:
\[
\mathcal{L}_{\mathrm{AMP}}^{-}(\psi)
=
\mathbb{E}_{(c,x_{\pi})\sim \rho_{\pi}}
\left[
\operatorname{BCE}
\left(l_{\theta,\psi}(c,x_{\pi}),0\right)
\right],
\]
where $\rho_{\pi}$ is the joint distribution of rollout contexts and policy
motion sequences. 

The adapter objective is
\[
\mathcal{L}_{\mathrm{CMP\text{-}AMP}}(\psi)
=
\mathcal{L}_{\mathrm{AMP}}^{+}(\psi)
+
\mathcal{L}_{\mathrm{AMP}}^{-}(\psi).
\]
The adapted motion-prior reward is
\[
r_{\mathrm{AMP}}(c,x)
=
-\log\!\left(
1-\sigma\!\left(l_{\theta,\psi}(c,x)\right)
\right).
\]

\paragraph{Extension to SMP.}
We apply CMP to SMP's pretrained denoiser. Given a reference motion
$x_E$, let $\tilde{x}_t$ denote its noisy version at diffusion step $t$.
CMP predicts
\[
\hat{\epsilon}_{\theta,\psi}(c,\tilde{x}_t,t)
=
\operatorname{sg}\!\left[\epsilon_{\theta}(\tilde{x}_t,t)\right]
+
\lambda_{\mathrm{res}}
\Delta\epsilon_{\psi}(c,\tilde{x}_t,t).
\]
The per-motion adapter loss is
\[
\mathcal{E}_{\psi}(c,x_E)
=
\mathbb{E}_{t,\epsilon}
\left[
\left\|
\epsilon-
\hat{\epsilon}_{\theta,\psi}(c,\tilde{x}_t,t)
\right\|_2^2
\right].
\]
For each denoising sample, this instantiates the generic adapter with
$z_{\theta}=\epsilon_{\theta}$,
$\Delta_{\psi}=\Delta\epsilon_{\psi}$, and
$y=\epsilon$. The relevance-weighted objective is
\[
\mathcal{L}_{\mathrm{CMP\text{-}SMP}}(\psi)
=
\mathbb{E}_{\substack{c\sim d_{\pi}\\x_E\sim p_E}}
\left[
\bar{w}_{\phi}(c,x_E)
\mathcal{E}_{\psi}(c,x_E)
\right].
\]
Only the adapter is updated, while the pretrained denoiser remains frozen.
During policy training, $\hat{\epsilon}_{\theta,\psi}$ replaces
$\epsilon_{\theta}$ in the original SMP reward, yielding
context-conditioned motion-prior guidance. The original SMP formulation
and full extension, including regularization and reward construction, are
detailed in Appendices~C
and~D, respectively. 

\section{Experiments}

We evaluate CMP in terms of task performance, sample efficiency,
context--motion compatibility, component contributions, and robustness to
reference dataset imbalance. We first compare CMP with the corresponding
base priors in return and sample efficiency
(Section~\ref{subsec:main}). We then examine learned relevance weights and
retrieved motions to assess context--motion alignment
(Section~\ref{subsec:alignment}). Ablations isolate the effects of adapter
capacity and relevance-learning components
(Section~\ref{subsec:ablation}), followed by robustness tests under
increasingly imbalanced reference distributions
(Section~\ref{subsec:robustness}).

\subsection{Experimental Setup}

We evaluate CMP on five humanoid control tasks requiring distinct forms of
context-dependent motion selection. In \textbf{Target Location}, the humanoid
moves toward a target specified in its local coordinate frame.
\textbf{Steering} requires it to follow a commanded movement direction and
speed while maintaining a desired facing direction.
\textbf{Trajectory Following} requires continuous tracking of a time-varying
path represented by future waypoints. In \textbf{Dodgeball}, the humanoid
reacts to the relative position and velocity of an incoming ball to avoid
collision. \textbf{Dribbling} further combines locomotion and object control,
requiring the humanoid to move a ball along a commanded direction at a target
speed.
All methods use the same locomotion reference set used by AMP
and distributed with MimicKit ~\citep{peng2021amp,peng2025mimickit}, without task-specific demonstrations or annotations.
Further task, dataset, and training details are provided in the appendix,
with the exact clip manifest included in the code.

\begin{figure*}[t]
    \centering
    \includegraphics[width=\textwidth]{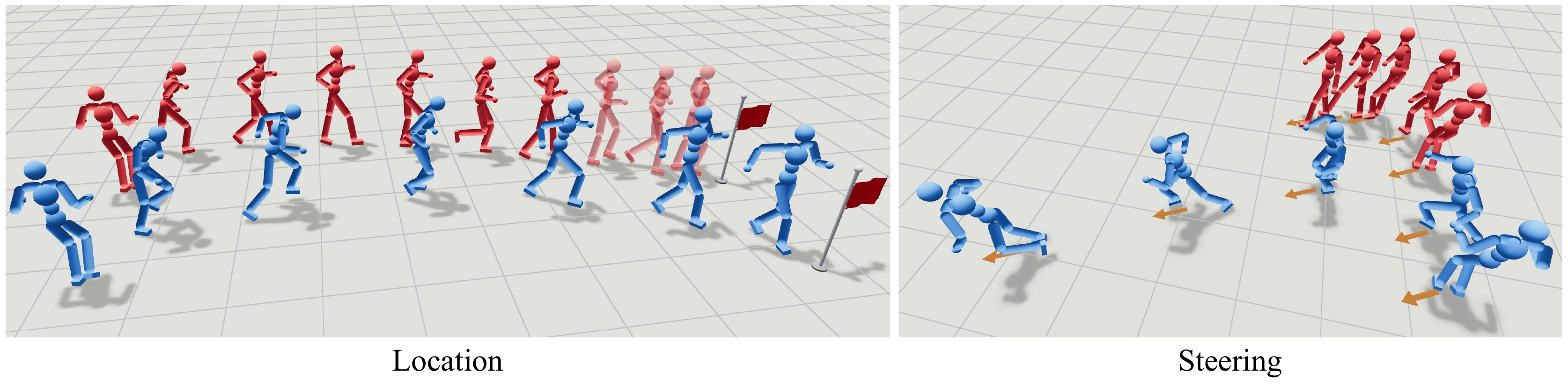}
    \caption{\textbf{Qualitative comparison of AMP and CMP-AMP.}
Temporally aligned snapshots compare CMP-AMP (blue) with AMP (red) on
Target Location and Steering. CMP-AMP reaches the target three intervals
earlier on Target Location and begins turning and accelerating earlier on
Steering. Additional visual comparisons are provided in the supplementary
material.}
    \label{fig:task}
\end{figure*}

\subsection{Main Results: Performance and Efficiency}
\label{subsec:main}

We first evaluate CMP with AMP as the primary instantiation and then examine
whether the same formulation generalizes to SMP.

\begin{figure*}[t]
\centering
\includegraphics[width=\textwidth]{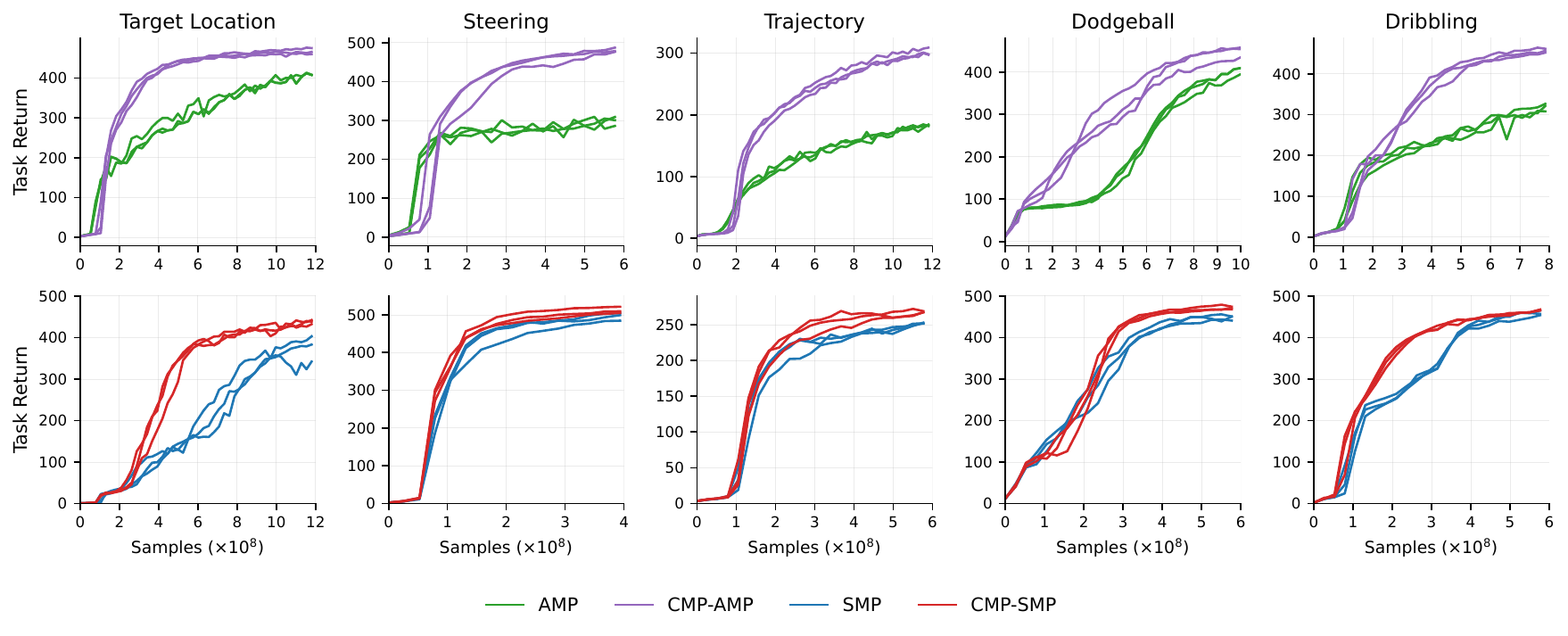}
\caption{\textbf{Learning curves across five humanoid control tasks.}
Top: CMP instantiated with AMP. Bottom: CMP extended to SMP.
All methods use the same task settings and training budgets.}
\label{fig:main_experiments_5x2}
\end{figure*}

\paragraph{Primary Results with AMP.}
Figure~\ref{fig:task} first provides temporally aligned qualitative
comparisons on Target Location and Steering. CMP-AMP reaches the target
earlier in Target Location and initiates turning and acceleration earlier
in Steering, indicating more efficient task execution.
Figure~\ref{fig:main_experiments_5x2} then compares AMP and CMP-AMP
quantitatively across five humanoid control tasks. CMP-AMP achieves higher task returns on all five tasks and reaches the
performance threshold with fewer environment interactions on four tasks,
while matching AMP on Steering.
Together, these results show that adapting the reference distribution to
the current task context improves both task performance and sample
efficiency.

\paragraph{Extension to Score-Matching Priors.}
Applying CMP to SMP yields consistent improvements in both final return and
sample efficiency across all five tasks, although the gains in final
performance are generally smaller than those obtained with AMP. This shows
that the relevance-weighted adaptation mechanism transfers beyond
adversarial motion priors, while the magnitude of improvement depends on
the underlying prior formulation.

\paragraph{Quantitative Comparison.}
Table~\ref{tab:main_results_combined} reports the test return and the
number of environment samples required to reach $80\%$ of the corresponding
base method's mean final return. The quantitative results corroborate the trends shown in Figure~\ref{fig:main_experiments_5x2}: CMP achieves its largest gains with AMP while demonstrating consistent transferability to SMP. Additional experiments on the G1 humanoid further validate these findings, as CMP-AMP improves both final return and sample efficiency across all five tasks (see Appendix~E).

\begin{table}[t]
\centering
\caption{\textbf{Main results.}
Results are mean $\pm$ standard deviation over three seeds.
Samples denote environment interactions ($\times 10^8$) required to reach
80\% of the corresponding base prior's mean final return.
Bold marks the better result within each prior pair; ties are both bold.}
\label{tab:main_results_combined}

\small
\renewcommand{\arraystretch}{1.05}
\setlength{\tabcolsep}{2.7pt}

\begin{tabular}{@{}lcc@{\hspace{5pt}}cc@{}}
\toprule
& \multicolumn{2}{c}{\textbf{AMP-based}}
& \multicolumn{2}{c}{\textbf{SMP-based}} \\
\cmidrule(lr){2-3}
\cmidrule(lr){4-5}
\textbf{Task}
& \textbf{AMP}
& \textbf{CMP-AMP}
& \textbf{SMP}
& \textbf{CMP-SMP} \\
\midrule

\multicolumn{5}{@{}l}{\textit{Test return} $\uparrow$} \\
\addlinespace[1pt]
Location
& $408 \pm 1$
& $\mathbf{467 \pm 8}$
& $376 \pm 31$
& $\mathbf{437 \pm 5}$ \\

Steering
& $299 \pm 12$
& $\mathbf{480 \pm 6}$
& $496 \pm 10$
& $\mathbf{512 \pm 8}$ \\

Trajectory
& $184 \pm 1$
& $\mathbf{302 \pm 6}$
& $252 \pm 1$
& $\mathbf{268 \pm 1}$ \\

Dodgeball
& $424 \pm 2$
& $\mathbf{458 \pm 12}$
& $447 \pm 6$
& $\mathbf{471 \pm 3}$ \\

Dribbling
& $319 \pm 10$
& $\mathbf{456 \pm 5}$
& $458 \pm 5$
& $\mathbf{466 \pm 2}$ \\

\midrule

\multicolumn{5}{@{}l}{%
    \textit{Samples} $(\times 10^8)$ $\downarrow$%
} \\
\addlinespace[1pt]
Location
& $6.5 \pm 0.8$
& $\mathbf{2.4 \pm 0.2}$
& $8.4 \pm 0.5$
& $\mathbf{4.7 \pm 0.5}$ \\

Steering
& $\mathbf{1.2 \pm 0.1}$
& $\mathbf{1.2 \pm 0.1}$
& $1.4 \pm 0.2$
& $\mathbf{1.3 \pm 0.1}$ \\

Trajectory
& $6.8 \pm 0.5$
& $\mathbf{2.7 \pm 0.2}$
& $2.2 \pm 0.2$
& $\mathbf{1.9 \pm 0.2}$ \\

Dodgeball
& $7.6 \pm 0.5$
& $\mathbf{5.6 \pm 0.7}$
& $3.1 \pm 0.2$
& $\mathbf{2.6 \pm 0.1}$ \\

Dribbling
& $5.2 \pm 0.3$
& $\mathbf{2.8 \pm 0.2}$
& $3.4 \pm 0.1$
& $\mathbf{2.2 \pm 0.2}$ \\

\bottomrule
\end{tabular}
\end{table}

\subsection{Analyzing Context--Motion Compatibility}
\label{subsec:alignment}

We next examine whether the contrastive relevance model captures meaningful
compatibility between task contexts and reference motion clips. We analyze
this alignment quantitatively through the learned relevance weights and
qualitatively through nearest-neighbor motion retrieval under different task
contexts.

\paragraph{Evaluation Contexts.}
We conduct this analysis on the Target Location task, where the context is
specified by a target position in the humanoid's local frame. We select seven
representative contexts covering different target directions and distances:
front-near (\textbf{FN}), front-medium (\textbf{FM}), front-far
(\textbf{FF}), side-medium (\textbf{SM}), side-far (\textbf{SF}),
back-medium (\textbf{BM}), and back-far (\textbf{BF}).

\begin{figure}[t]
\centering
\includegraphics[width=\linewidth]{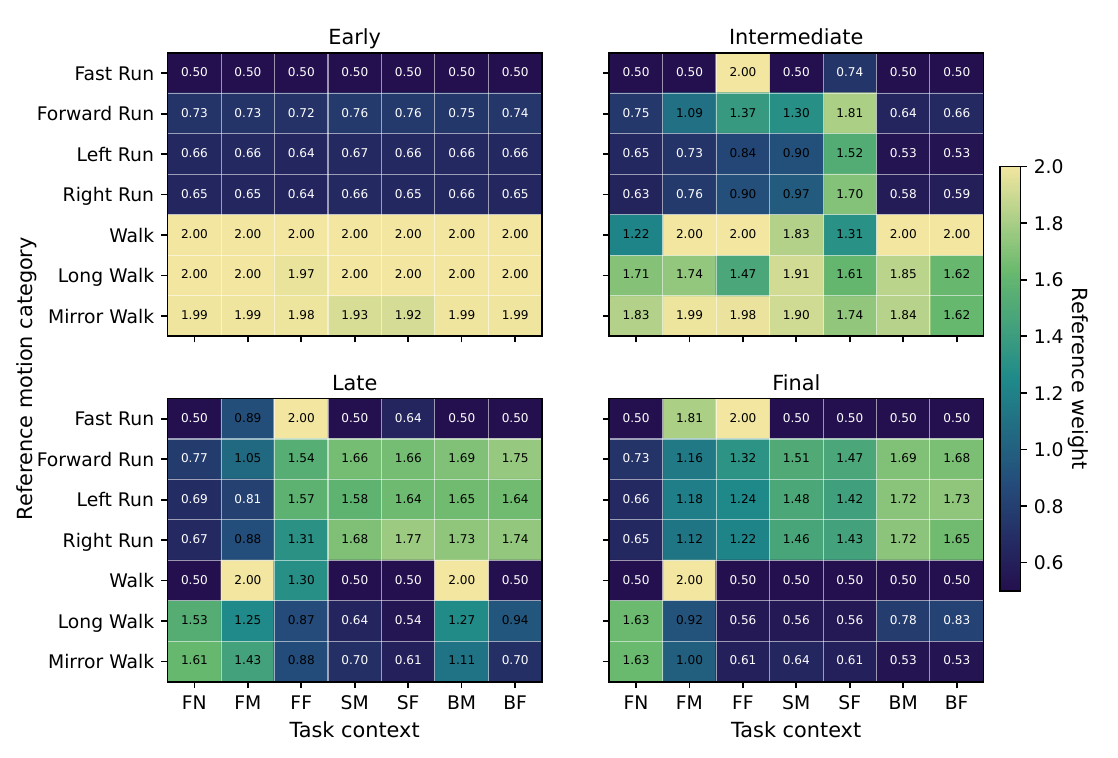}
\caption{\textbf{Evolution of learned relevance weights.}
Rows represent source motion groups and columns represent task contexts.
Group names are used only for visualization and are unavailable to CMP
during training. Early, Intermediate, Late, and Final correspond to
approximately $1.3$, $2.6$, $3.9$, and $5.2\times10^8$ environment
interactions, respectively.}
\label{fig:heatmap}
\end{figure}

\paragraph{Evolution of Relevance Weights.}
Figure~\ref{fig:heatmap} shows how the learned relevance weights evolve
across task contexts and training stages. Reference clips are grouped by
their source motion names, and each cell reports the mean weight within the
corresponding group.

Early in training, walking motions receive high weights across
contexts, consistent with the policy initially acquiring balance and basic
locomotion. The weights become increasingly context dependent as training
progresses.
At the final stage, \textbf{FN} favors long walking, whereas
\textbf{FF} favors fast running; side and back contexts instead emphasize
directional running motions. These patterns indicate that the relevance
model associates spatial task demands with suitable reference motions and
progressively adapts the prior as the policy evolves.

\paragraph{Motion Retrieval with Demonstration Support.}
\label{para:retrieval}

We further evaluate the learned alignment through nearest-neighbor motion
retrieval. For each representative context, reference clips are ranked by
the learned context--motion similarity. We compare full CMP with a variant
without the demonstration-positive loss
$\mathcal{L}_{\mathrm{demo}}$ to examine whether this branch anchors context
queries to the support of the reference dataset.

\begin{figure}[t]
\centering
\includegraphics[width=\linewidth]{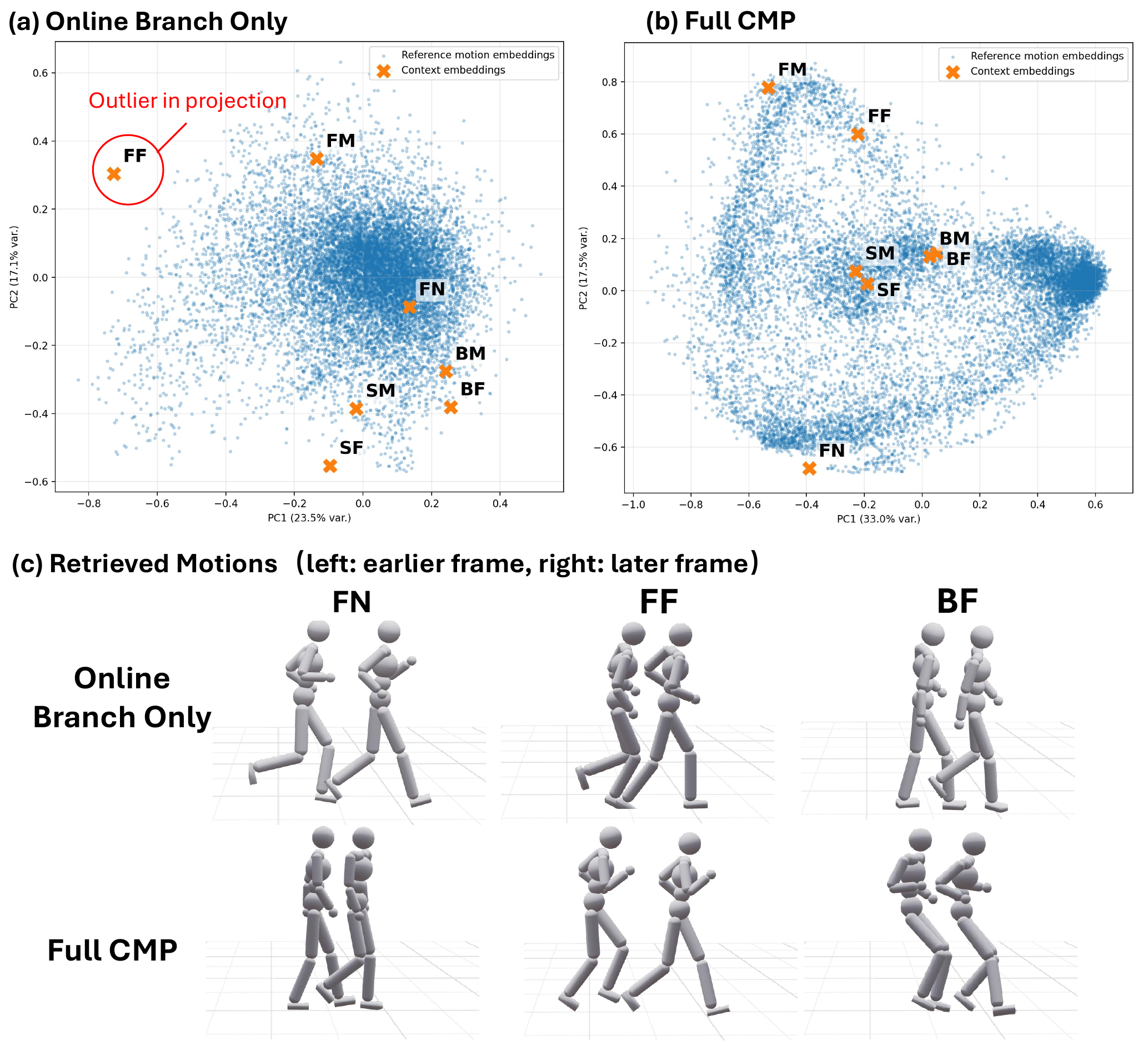}
\caption{\textbf{Demonstration support improves motion retrieval.}
Top: PCA projections of reference and context embeddings without
$\mathcal{L}_{\mathrm{demo}}$ and with the full CMP objective.
Bottom: the highest-ranked reference clips for three representative
contexts, with two frames shown per clip.}
\label{fig:representation}
\end{figure}

Without $\mathcal{L}_{\mathrm{demo}}$, context embeddings are shaped
primarily by online rollouts and may drift away from the reference support.
For example, the \textbf{FF} query appears separated from the reference
embeddings in the PCA projection in Fig.~\ref{fig:representation}.
The retrieved motions are also less context appropriate:
\textbf{FN} retrieves an unnecessarily dynamic motion, while
\textbf{BF} retrieves generic walking. With the full objective,
\textbf{FN}, \textbf{FF}, and \textbf{BF} retrieve walking, running, and
braking motions, respectively. These qualitative results suggest that
$\mathcal{L}_{\mathrm{demo}}$ anchors context queries to available
demonstrations and improves the reliability of context-conditioned motion
selection.

\subsection{Ablations of Contrastive Relevance}
\label{subsec:ablation}
We ablate three components of CMP-AMP: the context--conditioned adapter,
context--motion alignment, and the demonstration-positive branch, using AMP
as the unconditioned baseline. Corresponding CMP-SMP results are reported
in the supplementary material.

\paragraph{Ablation Variants.}
We compare full CMP-AMP with three variants.
\textit{Uniform Adapter} retains the same context-conditioned adapter but
sets all relevance weights to one, isolating architectural effects from
learned reference reweighting.
\textit{Online Branch Only} removes the demonstration-positive branch while
retaining the online contrastive objective; its effect is visualized and analyzed in
Fig.~\ref{fig:representation}.
\textit{Shuffled Relevance} randomly permutes the predicted relevance weights
across reference motions within each batch before they weight the
reference-side adapter loss. This preserves the batch-wise weight distribution
while disrupting its correspondence with the task contexts.

\begin{figure}[t]
    \centering
    \includegraphics[width=0.48\textwidth]
    {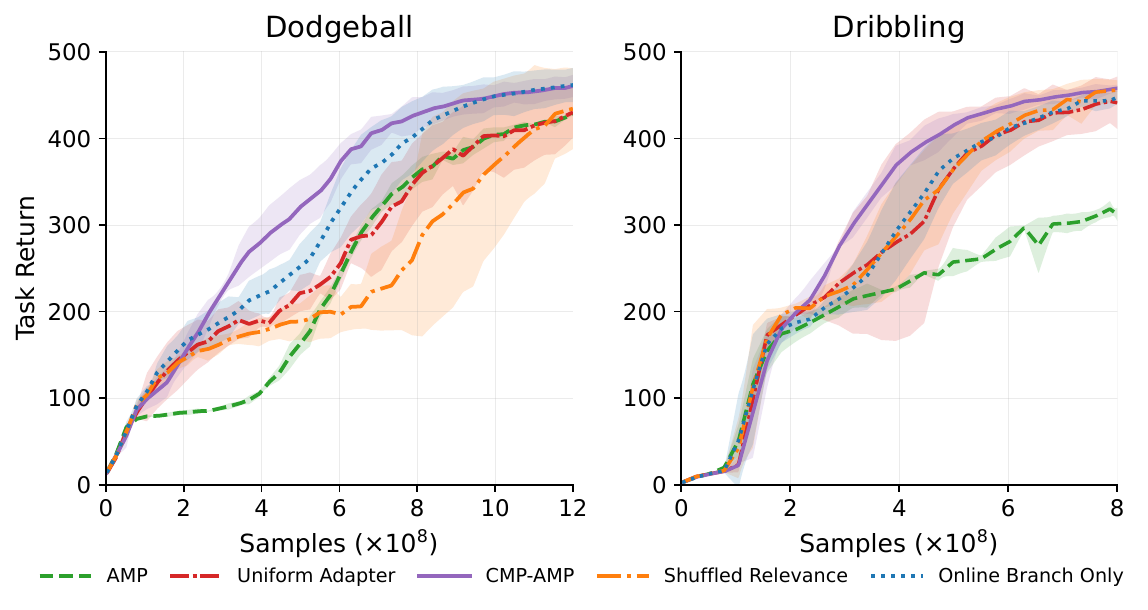}
    \caption{\textbf{CMP-AMP ablations.}
    Mean task return with standard deviation over three seeds on
    Dodgeball and Dribbling.}
    \label{fig:contrastive_ablation}
\end{figure}

\paragraph{Results.}
CMP-AMP reaches the threshold with the fewest samples. Uniform Adapter
outperforms AMP on Dribbling but provides little benefit on Dodgeball,
indicating that context-conditioned adapter capacity alone does not
consistently explain the gains delivered by CMP.

Shuffled Relevance retains a competitive final return on Dribbling but
learns more slowly than CMP-AMP. On Dodgeball, it exhibits substantially
higher variance and lower sample efficiency. Thus, meaningful
context--motion correspondence is particularly important for efficient and
reliable adaptation, even when shuffled weights can eventually achieve a
strong policy in some tasks.

Online Branch Only also reaches competitive final returns, but requires more
samples on both tasks. Together with the
retrieval analysis in Fig.~\ref{fig:representation}, these results support
the role of demonstration support in stabilizing relevance learning and
accelerating policy optimization.
For clarity, the main text focuses on the AMP-based instantiation.
Detailed numerical results for these experiments are reported in
Appendix~F, while the corresponding
ablation experiments for the SMP-based extension are provided in
Appendix~G.

\subsection{Reference Dataset Imbalance}
\label{subsec:robustness}

Finally, we evaluate the robustness of CMP-AMP to increasingly imbalanced
reference distributions on the Target Location task. Starting from the
original dataset, we increase the multiplicity of each walking motion by
factors of $2$, $5$, $20$, and $100$, while leaving all non-walking motions
unchanged. This modifies the empirical sampling frequencies without adding
or removing any motion category, producing reference distributions with increasingly high walking frequencies.

\begin{table}[t]
    \centering
    \caption{\textbf{Robustness to walking motion imbalance.}
    Parentheses denote relative changes from $\times 1$.}
    \label{tab:imbalance_robustness}
    \small
    \setlength{\tabcolsep}{4.5pt}
    \renewcommand{\arraystretch}{1.08}
    \begin{tabular}{@{}lcc@{}}
        \toprule
        Walking Factor
        & AMP Return $\uparrow$
      & CMP-AMP Return $\uparrow$ \\
        \midrule
        $\times 1$
        & $408$
        & $\mathbf{467}$ \\
        $\times 2$
        & $395\;(-3.2\%)$
        & $\mathbf{461\;(-1.3\%)}$ \\
        $\times 5$
        & $382\;(-6.4\%)$
        & $\mathbf{462\;(-1.1\%)}$ \\
        $\times 20$
        & $361\;(-11.5\%)$
        & $\mathbf{458\;(-1.9\%)}$ \\
        $\times 100$
        & $361\;(-11.5\%)$
        & $\mathbf{454\;(-2.8\%)}$ \\
        \bottomrule
    \end{tabular}
\end{table}

\paragraph{Results.}
As the walking proportion increases, AMP's return drops by up to $11.5\%$,
whereas CMP-AMP remains stable, decreasing by at most $2.8\%$. Under the most
severe imbalance, CMP-AMP retains a return of $454$ versus $361$ for AMP.
These results show that CMP reduces the sensitivity of task-agnostic priors
to dominant motion frequencies by emphasizing context-compatible references.
The corresponding learning curves are provided in Appendix~H.

\section{Conclusion and Limitations}

We introduced Context-Aware Motion Priors (CMP), which learns
context--motion relevance from online policy experience to adapt the effective
reference distribution of a motion prior. Across five humanoid control tasks,
CMP substantially improves AMP's performance and sample efficiency and yields
smaller but consistent gains with SMP. The learned weights and retrieved
motions show meaningful context--motion alignment, and CMP remains robust to
severe imbalance in reference frequencies. These results suggest that motion
priors can adapt during downstream learning to emphasize reference motions
relevant to the current context.

CMP remains limited by the support of the reference dataset and cannot recover
task-relevant behaviors that are absent. Its reliance on policy advantages
also makes relevance learning sensitive to critic error and insufficient
exploration. Our evaluation is limited to structured contexts and simulated
humanoids; high-dimensional perception, real-world deployment, and direct
evaluation of motion quality remain future work. Finally, the smaller gains
with SMP indicate that CMP's effectiveness also depends on how the underlying
prior represents and translates motion supervision into policy guidance.

\newpage

\bibliography{aaai2027}

\newpage

\appendix

\section{Adapter Loss and Reweighting Equivalence}
\label{app:proof_reweighting}

\paragraph{Connection to the Main Objective.}
Section~3.1 defines CMP using a relevance-weighted
adapter objective, while leaving the per-sample adapter loss
$\ell_{\psi}^{\mathrm{adapt}}(c,x;\theta)$ abstract. Here, we first detail
how this loss is constructed by applying a context-conditioned residual
adapter to the detached output of the base motion prior. We then show that,
under the ideal context-wise normalized weights, the weighted objective is
equivalent to training with reference motions drawn from
$q_{\phi}(x\mid c)$. The minibatch normalization and clipping used in
practice provide a bounded approximation to this ideal objective.

\paragraph{Context-Conditioned Adapter Objective.}
Let $z_{\theta}(x)$ denote the output produced by the base motion prior
for reference motion $x$, and let $y(x)$ denote the corresponding training
target. We write $\mathcal{D}(\cdot,\cdot)$ for the prior-specific loss
function. During adapter optimization, the base prediction is treated as a
stop-gradient input:
\[
\bar{z}_{\theta}(x)
=
\operatorname{sg}\!\left[z_{\theta}(x)\right],
\]
where $\operatorname{sg}(\cdot)$ denotes the stop-gradient operation. CMP
adds a context-conditioned residual adapter $\Delta_{\psi}(c,x)$ and forms
\[
z_{\theta,\psi}(c,x)
=
\bar{z}_{\theta}(x)
+
\lambda_{\mathrm{res}}\Delta_{\psi}(c,x),
\]
where $\psi$ denotes the adapter parameters and $\lambda_{\mathrm{res}}$
controls the residual strength. The reference-side adapter loss is
\[
\ell_{\psi}^{\mathrm{adapt}}(c,x;\theta)
=
\mathcal{D}
\left(
z_{\theta,\psi}(c,x),
y(x)
\right).
\]
The dependence on $\theta$ is numerical only during adapter optimization:
the base prior may still be updated separately through its original
objective. We instantiate $\mathcal{D}$, $z_{\theta}$, and $y$ for AMP and
SMP in Section~3.3.

For the ideal population-level weights, define
\[
\mathcal{L}_{\mathrm{adapt}}^{\mathrm{ideal}}
(\psi\mid\theta,\phi)
=
\mathbb{E}_{\substack{c\sim d_{\pi}\\x\sim p_E}}
\left[
\operatorname{sg}\!\left[w_{\phi}(c,x)\right]
\ell_{\psi}^{\mathrm{adapt}}(c,x;\theta)
\right].
\]
This objective updates only the adapter parameters $\psi$. Since the
stop-gradient operation does not change the numerical values of the
weights, it does not affect the reweighting equivalence derived below.

\paragraph{Reweighting Equivalence.}
Recall from Section~3.1 that
\[
q_{\phi}(x\mid c)
=
w_{\phi}(c,x)p_E(x),
\qquad
\mathbb{E}_{x\sim p_E}
\left[
w_{\phi}(c,x)
\right]
=
1.
\]
The normalization condition ensures that
$q_{\phi}(\cdot\mid c)$ is a valid distribution. For any fixed context
$c$, we have
\[
\begin{aligned}
&
\mathbb{E}_{x\sim q_{\phi}(\cdot\mid c)}
\left[
\ell_{\psi}^{\mathrm{adapt}}(c,x;\theta)
\right]
\\
&=
\int_{\mathcal X}
q_{\phi}(x\mid c)
\ell_{\psi}^{\mathrm{adapt}}(c,x;\theta)
\,\mathrm{d}x
\\
&=
\int_{\mathcal X}
p_E(x)w_{\phi}(c,x)
\ell_{\psi}^{\mathrm{adapt}}(c,x;\theta)
\,\mathrm{d}x
\\
&=
\mathbb{E}_{x\sim p_E}
\left[
w_{\phi}(c,x)
\ell_{\psi}^{\mathrm{adapt}}(c,x;\theta)
\right].
\end{aligned}
\]
Taking the expectation over contexts $c\sim d_{\pi}$ yields
\[
\begin{aligned}
&
\mathbb{E}_{\substack{c\sim d_{\pi}\\
x\sim q_{\phi}(\cdot\mid c)}}
\left[
\ell_{\psi}^{\mathrm{adapt}}(c,x;\theta)
\right]
\\
&=
\mathbb{E}_{\substack{c\sim d_{\pi}\\
x\sim p_E}}
\left[
w_{\phi}(c,x)
\ell_{\psi}^{\mathrm{adapt}}(c,x;\theta)
\right].
\end{aligned}
\]
Since stop gradient does not alter the numerical value of the weights,
the same identity applies to the adapter update using
$\operatorname{sg}[w_{\phi}(c,x)]$. This establishes that, under the ideal
normalized weights, relevance-weighted training with samples from $p_E$
is equivalent to training with reference motions drawn from the effective
distribution $q_{\phi}(x\mid c)$.

For an empirical reference dataset
$\mathcal{D}_E=\{x_n\}_{n=1}^{N}$,
\[
p_E(x)
=
\frac{1}{N}
\sum_{n=1}^{N}
\delta_{x_n}(x),
\]
where $\delta_{x_n}$ denotes a point mass at $x_n$. The corresponding
effective probability assigned to $x_n$ is
\[
q_{\phi}(x_n\mid c)
=
\frac{1}{N}w_{\phi}(c,x_n),
\]
with
\[
\frac{1}{N}
\sum_{n=1}^{N}
w_{\phi}(c,x_n)
=
1.
\]
Hence, for a fixed context,
\[
\begin{aligned}
&
\mathbb{E}_{x\sim q_{\phi}(\cdot\mid c)}
\left[
\ell_{\psi}^{\mathrm{adapt}}(c,x;\theta)
\right]
\\
&\quad=
\frac{1}{N}
\sum_{n=1}^{N}
w_{\phi}(c,x_n)\,
\ell_{\psi}^{\mathrm{adapt}}(c,x_n;\theta).
\end{aligned}
\]
This is the empirical form of the ideal context-wise reweighting
objective. In practice, CMP replaces $w_{\phi}$ with the detached
minibatch-normalized and clipped weight $\bar{w}_{\phi}$ defined in
Section~3.1. This yields an efficient bounded
approximation without normalizing over the full reference dataset for
every context.

\section{Original Adversarial Motion Prior}
\label{app:amp_original}

This section briefly reviews the original Adversarial Motion Prior (AMP)
objective used as the base prior in CMP-AMP. AMP learns a discriminator that distinguishes reference motion samples from policy-generated motion samples. Let $x$ denote a motion observation, such as a short history of humanoid poses and velocities. Let $p_E(x)$ be the reference motion distribution induced by the
demonstration dataset, and let $p_{\pi}(x)$ be the motion distribution
induced by the current policy.

AMP parameterizes a discriminator $D_{\theta}(x)$ using a scalar logit $l_{\theta}(x)$:
$$
D_{\theta}(x)
=
\sigma(l_{\theta}(x)),
$$
where $\sigma(\cdot)$ is the sigmoid function. The discriminator is trained with a binary classification objective that assigns label $1$ to reference motions and label $0$ to policy motions:
$$
\begin{aligned}
\mathcal{L}_{D}^{\mathrm{AMP}}
=&\;
\mathbb{E}_{p_E}
\left[
\mathcal{L}_{\mathrm{BCE}}(l_{\theta}(x_E),1)
\right] \\
&+
\mathbb{E}_{p_{\pi}}
\left[
\mathcal{L}_{\mathrm{BCE}}(l_{\theta}(x_{\pi}),0)
\right].
\end{aligned}
$$
where the binary cross-entropy loss is defined as
$$
\mathcal{L}_{\mathrm{BCE}}(l,y)
=
-y \log \sigma(l)
-
(1-y)\log(1-\sigma(l)).
$$

Equivalently, the positive and negative parts of the AMP discriminator loss can be written as
$$
\mathcal{L}_{D}^{\mathrm{pos}}
=
\mathbb{E}_{x_E \sim p_E}
\left[
-\log D_{\theta}(x_E)
\right],
$$
and
$$
\mathcal{L}_{D}^{\mathrm{neg}}
=
\mathbb{E}_{x_{\pi} \sim p_{\pi}}
\left[
-\log(1-D_{\theta}(x_{\pi}))
\right].
$$
The full discriminator objective is therefore
$$
\mathcal{L}_{D}^{\mathrm{AMP}}
=
\mathcal{L}_{D}^{\mathrm{pos}}
+
\mathcal{L}_{D}^{\mathrm{neg}}.
$$

During policy optimization, AMP converts the discriminator output into a style reward:
$$
r_{\mathrm{AMP}}(x)
=
-\log\left(1-D_{\theta}(x)\right)
=
-\log\left(1-\sigma(l_{\theta}(x))\right).
$$
This reward encourages the policy to generate motions that the discriminator classifies as reference-like. In practice, AMP implementations may include additional regularization terms for stabilizing discriminator training, such as gradient penalties or logit regularization. These terms are orthogonal to our formulation and are omitted here. In CMP-AMP, the base discriminator parameters $\theta$ continue to be
updated by this original objective, while gradients from the adapter
objective do not propagate to $\theta$.

\section{Original Score-Matching Motion Prior}
\label{app:smp_original}

This section reviews the original Score-Matching Motion Prior (SMP)~\citep{mu2025smp}
formulation used as the base prior in CMP-SMP.
SMP represents a task-agnostic motion prior using a pretrained diffusion
model; we use its unconditional instantiation. Let $x_0 \sim p_E$ denote a clean reference motion sample. The forward diffusion process perturbs $x_0$ into a noisy sample
$\tilde{x}_t$:
\[
\tilde{x}_t
=
\sqrt{\bar{\alpha}_t}x_0
+
\sqrt{1-\bar{\alpha}_t}\epsilon,
\]
where $t\sim\mathcal{U}\{0,\ldots,T-1\}$,
$\epsilon\sim\mathcal{N}(0,I)$, and $\bar{\alpha}_t$ is the cumulative
noise schedule.

The denoising model $\epsilon_{\theta}(\tilde{x}_t,t)$ is trained to predict the injected Gaussian noise. The standard SMP training objective is
$$
\mathcal{L}_{\mathrm{SMP}}(\theta)
=
\mathbb{E}_{x_0 \sim p_E,\, t,\, \epsilon}
\left[
\left\|
\epsilon
-
\epsilon_{\theta}(\tilde{x}_t,t)
\right\|^2
\right].
$$
This objective learns an unconditional motion prior from the reference dataset. During policy optimization, SMP evaluates policy-generated motions using
ensemble score matching. Let $x$ denote a policy motion and let
$\mathcal{K}$ be a fixed set of diffusion timesteps. For each
$k\in\mathcal{K}$, independent Gaussian noise is applied:
\[
\tilde{x}_{k}
=
\sqrt{\bar{\alpha}_{k}}\,x
+
\sqrt{1-\bar{\alpha}_{k}}\,\epsilon_{k},
\qquad
\epsilon_{k}\sim\mathcal{N}(0,I).
\]
The per-timestep denoising error is
\[
e_{\theta,k}(x)
=
\frac{1}{d_x}
\left\|
\epsilon_{k}
-
\epsilon_{\theta}(\tilde{x}_{k},k)
\right\|_2^2,
\]
where $d_x$ is the motion-feature dimension. To account for differences
in error scale across noise levels, SMP normalizes each error by its
running mean $\mu_k$:
\[
\tilde{e}_{\theta,k}(x)
=
\frac{e_{\theta,k}(x)}
{\max(\mu_k,\epsilon_{\mathrm{norm}})}.
\]
The SMP reward is then
\[
r_{\mathrm{SMP}}(x)
=
\exp\!\left(
-\frac{\kappa}{|\mathcal{K}|}
\sum_{k\in\mathcal{K}}
\tilde{e}_{\theta,k}(x)
\right),
\]
where $\kappa>0$ controls the reward scale. Following SMP, we use
$\mathcal{K}=\{22,15,8\}$.

CMP-SMP retains the pretrained denoiser as a frozen base and learns a
context-conditioned residual adapter from relevance-weighted reference
motions. Appendix~\ref{app:cmp_smp} details the adapter objective and
context-conditioned reward.

\section{CMP Extension to Score-Matching Motion Priors}
\label{app:cmp_smp}

Let $\epsilon_{\theta}(\tilde{x}_t,t)$ denote the pretrained SMP denoiser, which
predicts the noise added to a motion sequence at diffusion timestep $t$.
For a clean reference motion $x_E\sim p_E$, the forward diffusion process is
\[
\tilde{x}_t
=
\sqrt{\bar{\alpha}_t}\,x_E
+
\sqrt{1-\bar{\alpha}_t}\,\epsilon,
\qquad
\epsilon\sim\mathcal{N}(0,I),
\]
where $\bar{\alpha}_t$ is determined by the diffusion noise schedule.

CMP introduces a context-conditioned residual denoising adapter
$\Delta\epsilon_{\psi}(c,\tilde{x}_t,t)$. During adapter optimization, the
prediction of the pretrained denoiser is treated as stop gradient:
\[
\hat{\epsilon}_{\theta,\psi}(c,\tilde{x}_t,t)
=
\operatorname{sg}\!\left[
\epsilon_{\theta}(\tilde{x}_t,t)
\right]
+
\lambda_{\mathrm{res}}
\Delta\epsilon_{\psi}(c,\tilde{x}_t,t),
\]
where $\lambda_{\mathrm{res}}$ controls the contribution of the residual
adapter. The pretrained SMP parameters $\theta$ remain unchanged.
For a reference motion, we define the per-motion denoising
error as
\[
\mathcal{E}_{\psi}(c,x_E)
=
\mathbb{E}_{t,\epsilon}
\left[
\left\|
\epsilon-
\hat{\epsilon}_{\theta,\psi}(c,\tilde{x}_t,t)
\right\|_2^2
\right].
\]
During adapter training, $t$ is sampled uniformly from
$\{0,\ldots,T-1\}$ and $\epsilon\sim\mathcal{N}(0,I)$.
Using the detached relevance weight
$\bar{w}_{\phi}(c,x_E)$ defined in
Section~3.1, the reference-side denoising loss is
\[
\mathcal{L}_{\mathrm{SMP}}^{\mathrm{adapt}}(\psi)
=
\mathbb{E}_{\substack{c\sim d_{\pi}\\x_E\sim p_E}}
\left[
\operatorname{sg}\!\left[\bar{w}_{\phi}(c,x_E)\right]
\frac{\mathcal{E}_{\psi}(c,x_E)}{d_x}
\right].
\]
Thus, the denoising adapter receives different effective reference
distributions under different task contexts, while gradients do not
propagate into either the relevance model or the pretrained denoiser.

We additionally regularize the residual prediction to prevent the adapter
from overriding the pretrained motion prior:
\[
\mathcal{L}_{\mathrm{res}}(\psi)
=
\mathbb{E}_{\substack{
c\sim d_{\pi},\,x_E\sim p_E\\
t\sim\mathcal{U}\{0,\ldots,T-1\}\\
\epsilon\sim\mathcal{N}(0,I)}}
\left[
\frac{1}{d_x}
\left\|
\lambda_{\mathrm{res}}
\Delta\epsilon_{\psi}(c,\tilde{x}_t,t)
\right\|_2^2
\right].
\]
The total CMP-SMP adapter objective is
\[
\mathcal{L}_{\mathrm{CMP\text{-}SMP}}^{\mathrm{total}}(\psi)
=
\mathcal{L}_{\mathrm{SMP}}^{\mathrm{adapt}}(\psi)
+
\eta\mathcal{L}_{\mathrm{res}}(\psi),
\]
where $\eta$ controls the residual regularization strength.

During policy optimization, let $x$ denote a policy-generated motion
sequence. CMP-SMP follows the ensemble score-matching construction of SMP.
For every $k\in\mathcal{K}$, we sample
$\epsilon_k\sim\mathcal{N}(0,I)$ and form
\[
\tilde{x}_{k}
=
\sqrt{\bar{\alpha}_{k}}\,x
+
\sqrt{1-\bar{\alpha}_{k}}\,\epsilon_k.
\]
The adapted per-timestep denoising error is
\[
e_{\theta,\psi,k}(c,x)
=
\frac{1}{d_x}
\left\|
\epsilon_k
-
\hat{\epsilon}_{\theta,\psi}(c,\tilde{x}_{k},k)
\right\|_2^2.
\]
Using the same timestep-wise normalization procedure as SMP, let
$\mu_k^{\mathrm{adapt}}$ denote the running mean of
$e_{\theta,\psi,k}$. We define
\[
\tilde{e}_{\theta,\psi,k}(c,x)
=
\frac{
e_{\theta,\psi,k}(c,x)
}{
\max(\mu_k^{\mathrm{adapt}},\epsilon_{\mathrm{norm}})
}.
\]
The context-conditioned motion-prior reward is
\[
r_{\mathrm{CMP\text{-}SMP}}(c,x)
=
\exp\!\left(
-\frac{\kappa}{|\mathcal{K}|}
\sum_{k\in\mathcal{K}}
\tilde{e}_{\theta,\psi,k}(c,x)
\right),
\]
where $\mathcal{K}=\{22,15,8\}$. The policy reward combines this adapted prior reward with the task reward:
\[
r
=
\lambda_{\mathrm{task}}r_{\mathrm{task}}
+
\lambda_{\mathrm{prior}}
r_{\mathrm{CMP\text{-}SMP}}(c,x),
\]
where $\lambda_{\mathrm{task}}$ and $\lambda_{\mathrm{prior}}$ weight the
task and motion-prior rewards, respectively.

\section{G1 Results}
\label{app:main_g1}

\begin{figure*}[t]
    \centering
    \includegraphics[width=2 \columnwidth]{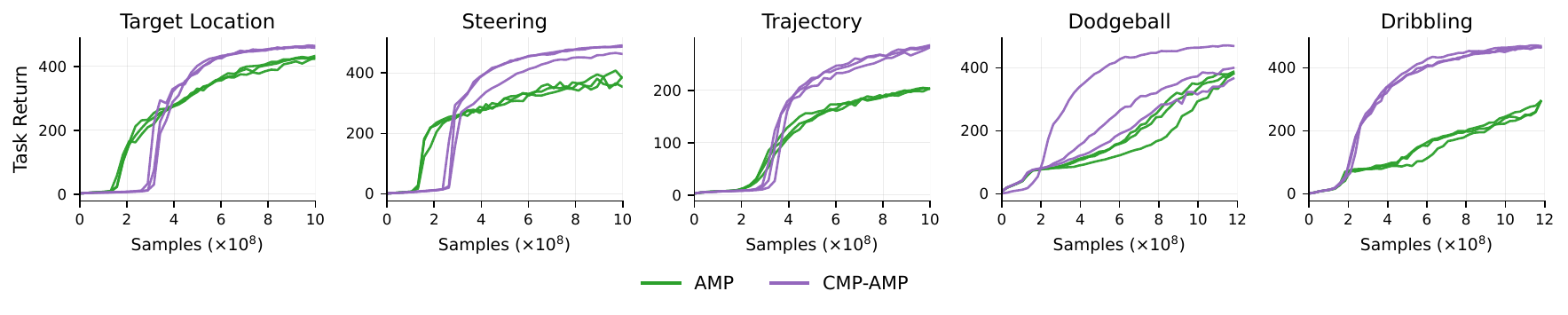}
    \caption{\textbf{G1 humanoid results.}
Learning curves of AMP and CMP-AMP on five humanoid control tasks.
Each curve corresponds to one of three random seeds.}
    \label{fig:g1_results}
\end{figure*}

To evaluate CMP on a different humanoid morphology, we further report
CMP-AMP results on a simulated G1 humanoid model.
Figure~\ref{fig:g1_results} shows the learning curves of AMP and CMP-AMP
across five tasks. Table~\ref{tab:g1_results} summarizes the final test
return and the number of environment samples required to reach $80\%$ of
AMP's mean final return. All results are averaged over three random seeds.

\begin{table}[t]
    \centering
    \caption{\textbf{G1 humanoid results.}
Mean and standard deviation over three seeds.
``Samples to 80\%'' denotes the number of environment samples
($\times 10^8$) required to reach $80\%$ of AMP's mean final return.}
    \label{tab:g1_results}
    \small
    \setlength{\tabcolsep}{3pt}
    \renewcommand{\arraystretch}{1.05}
    \begin{tabular}{lcc}
        \toprule
        Method
        & Task Return $\uparrow$
        & Samples to 80\% $\downarrow$ \\
        \midrule

        \multicolumn{3}{l}{\textit{Target Location}}\\
        AMP
        & $428 \pm 3$
        & $5.6 \pm 0.1$ \\
        CMP-AMP
        & $\mathbf{459 \pm 4}$
        & $\mathbf{4.5 \pm 0.1}$ \\

        \midrule

        \multicolumn{3}{l}{\textit{Steering}}\\
        AMP
        & $375 \pm 14$
        & $4.9 \pm 0.3$ \\
        CMP-AMP
        & $\mathbf{480 \pm 12}$
        & $\mathbf{3.4 \pm 0.4}$ \\

        \midrule

        \multicolumn{3}{l}{\textit{Trajectory Following}}\\
        AMP
        & $203 \pm 0$
        & $5.9 \pm 0.1$ \\
        CMP-AMP
        & $\mathbf{283 \pm 2}$
        & $\mathbf{4.0 \pm 0.1}$ \\

        \midrule

        \multicolumn{3}{l}{\textit{Dodgeball}}\\
        AMP
        & $384 \pm 3$
        & $9.6 \pm 0.9$ \\
        CMP-AMP
        & $\mathbf{412 \pm 43}$
        & $\mathbf{6.7 \pm 2.4}$ \\

        \midrule

        \multicolumn{3}{l}{\textit{Dribbling}}\\
        AMP
        & $294 \pm 0$
        & $10.1 \pm 0.4$ \\
        CMP-AMP
        & $\mathbf{467 \pm 2}$
        & $\mathbf{2.9 \pm 0.1}$ \\

        \bottomrule
    \end{tabular}
\end{table}

\section{Detailed CMP-AMP Ablation Results}
\label{app:contrastive_ablation}

Table~\ref{tab:contrastive_ablation} reports the detailed numerical results
for the CMP-AMP ablations presented in the main text. \textit{Uniform
Adapter} retains the context-conditioned adapter but assigns uniform
reference weights. \textit{Shuffled Relevance} permutes the predicted
weights across reference motions within each batch, while
\textit{Online Branch Only} removes the demonstration-positive branch.

\begin{table}[t]
    \centering
    \caption{\textbf{Contrastive relevance ablations.}
Final test return and environment samples $(\times 10^8)$ required to reach
$80\%$ of AMP's mean final return. Results are reported as mean and standard
deviation over three seeds. Bold indicates the best result; ties at the
reported precision are both bold.}
    \label{tab:contrastive_ablation}
    \small
    \setlength{\tabcolsep}{4pt}
    \renewcommand{\arraystretch}{1.08}
    \begin{tabular}{lcc}
        \toprule
        Method
        & Task Return $\uparrow$
        & Samples to $80\%$ $\downarrow$ \\
        \midrule

        \multicolumn{3}{l}{\textit{Dribbling}} \\
        AMP
        & $319 \pm 10$
        & $5.2 \pm 0.3$ \\
        Uniform Adapter
        & $441 \pm 30$
        & $3.7 \pm 1.3$ \\
        
        Shuffled Relevance
        & $453 \pm 7$
        & $3.8 \pm 0.2$ \\
        Online Branch Only
        & $447 \pm 8$
        & $3.6 \pm 0.5$ \\
        CMP-AMP
        & $\mathbf{456 \pm 5}$
        & $\mathbf{2.8 \pm 0.2}$ \\

        \midrule
        \multicolumn{3}{l}{\textit{Dodgeball}} \\
        AMP
        & $424 \pm 2$
        & $7.6 \pm 0.5$ \\
        Uniform Adapter
        & $430 \pm 29$
        & $7.8 \pm 0.9$ \\
        Shuffled Relevance
        & $434 \pm 47$
        & $9.5 \pm 1.7$ \\
        Online Branch Only
        & $\mathbf{458 \pm 19}$
        & $6.5 \pm 0.7$ \\
        CMP-AMP
        & $\mathbf{458 \pm 12}$
        & $\mathbf{5.6 \pm 0.7}$ \\

        \bottomrule
    \end{tabular}
\end{table}

CMP-AMP reaches the performance threshold with the fewest samples on both
tasks. The adapter alone improves Dribbling but provides little benefit on
Dodgeball, while shuffled relevance reduces efficiency and substantially
increases variance on Dodgeball. Removing the demonstration-positive branch
also slows learning, despite retaining competitive final performance.

\section{Ablations of the Score-Matching Extension}
\label{app:smp_ablation}

We repeat the same ablations for CMP-SMP to examine whether the contrastive
relevance components remain effective with a score-matching prior.

\begin{figure}[t]
    \centering
    \hspace*{-0.03\columnwidth}
    \includegraphics[width=0.485\textwidth]
    {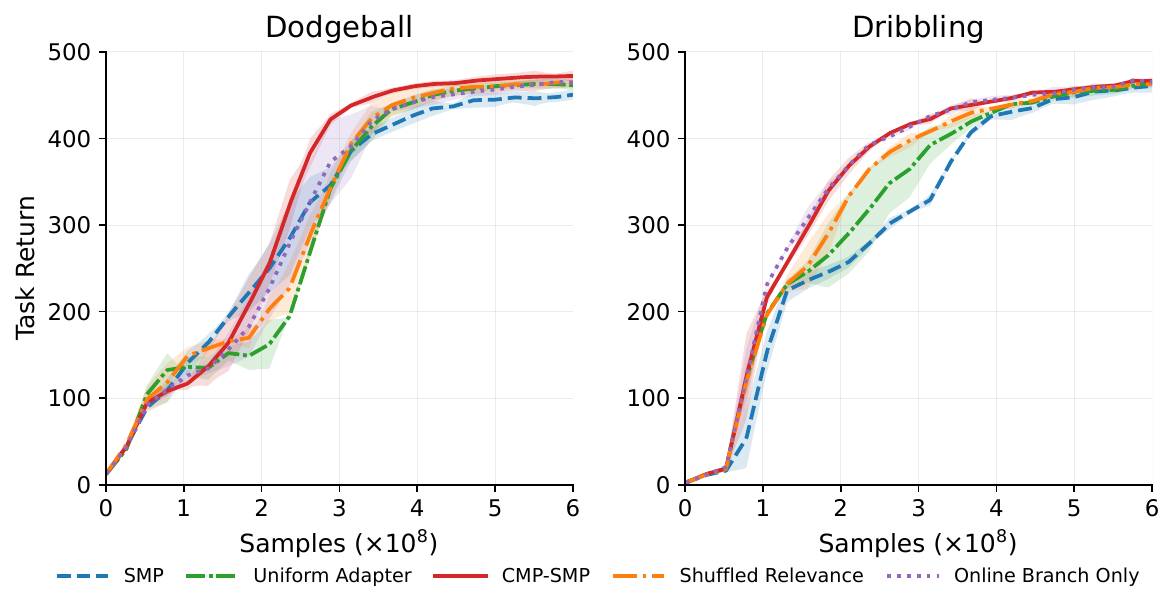}
    \caption{\textbf{CMP-SMP ablations.}
    Learning curves on Dodgeball and Dribbling. Solid curves show the mean
    task return and shaded regions indicate the standard deviation over
    three seeds.}
    \label{fig:smp_ablation}
\end{figure}

\begin{table}[t]
    \centering
    \caption{\textbf{Ablations of the score-matching extension.}
Final test return and environment samples $(\times 10^8)$ required to reach
$80\%$ of SMP's mean final return. Results are reported as mean and standard
deviation over three seeds. Bold indicates the best result; ties at the
reported precision are both bold.}
    \label{tab:smp_ablation}
    \small
    \setlength{\tabcolsep}{4pt}
    \renewcommand{\arraystretch}{1.08}
    \begin{tabular}{lcc}
        \toprule
        Method
        & Task Return $\uparrow$
        & Samples to $80\%$ $\downarrow$ \\
        \midrule

        \multicolumn{3}{l}{\textit{Dribbling}} \\
        SMP
        & $458 \pm 5$
        & $3.4 \pm 0.1$ \\
        Uniform Adapter
        & $461 \pm 6$
        & $3.0 \pm 0.2$ \\
        Shuffled Relevance
        & $463 \pm 2$
        & $2.4 \pm 0.1$ \\
        Online Branch Only
        & $465 \pm 6$
        & $\mathbf{2.2 \pm 0.1}$ \\
        CMP-SMP
        & $\mathbf{466 \pm 2}$
        & $\mathbf{2.2 \pm 0.1}$ \\

        \midrule
        \multicolumn{3}{l}{\textit{Dodgeball}} \\
        SMP
        & $447 \pm 6$
        & $3.1 \pm 0.2$ \\
        Uniform Adapter
        & $464 \pm 6$
        & $2.8 \pm 0.4$ \\
        Shuffled Relevance
        & $465 \pm 7$
        & $3.0 \pm 0.1$ \\
        Online Branch Only
        & $465 \pm 5$
        & $2.8 \pm 0.3$ \\
        CMP-SMP
        & $\mathbf{471 \pm 3}$
        & $\mathbf{2.6 \pm 0.1}$ \\

        \bottomrule
    \end{tabular}
\end{table}

CMP-SMP achieves the highest final return on both tasks. On Dodgeball, it
also reaches the performance threshold with the fewest samples. Uniform Adapter, Shuffled Relevance, and Online Branch Only match or improve
upon SMP in final return, but none matches the full method in both
performance and efficiency. This indicates that meaningful context--motion correspondence
and demonstration support remain beneficial when CMP is integrated with a
score-matching prior.

On Dribbling, CMP-SMP and Online Branch Only reach the threshold at the same
rate, while CMP-SMP achieves a slightly higher final return with lower
variance. Shuffled Relevance also improves efficiency over SMP but remains
slower than the variants that preserve learned context--motion
correspondence. Thus, the demonstration-positive branch has a smaller
effect on Dribbling, whereas the full relevance formulation provides clearer
benefits on Dodgeball.

\section{Learning Curves under Reference Dataset Imbalance}
\label{app:imbalance_curves}

\begin{figure}[t]
    \centering
    \includegraphics[width=\linewidth]{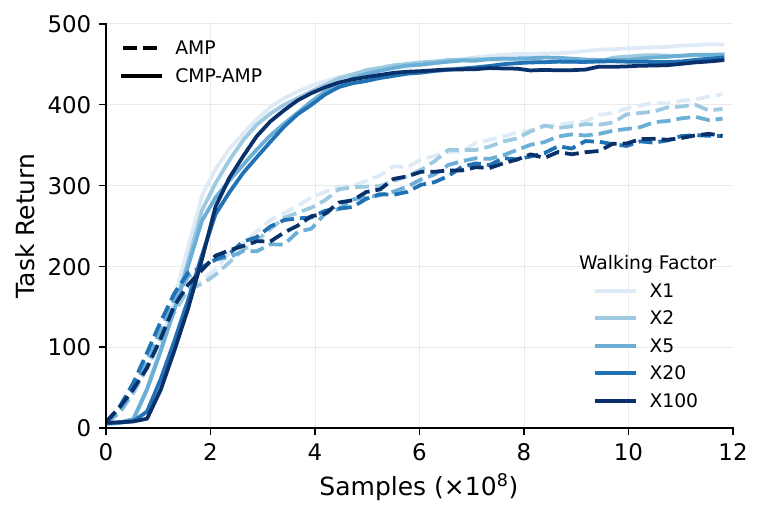}
    \caption{\textbf{Learning curves under walking motion imbalance.}
    Mean task return and standard deviation over three seeds under different
    walking multiplicities. CMP-AMP remains substantially more stable than
    AMP as the reference distribution becomes increasingly imbalanced.}
    \label{fig:imbalance_curves}
\end{figure}

Figure~\ref{fig:imbalance_curves} shows the learning curves corresponding
to the reference imbalance results reported in the main paper. We compare AMP and CMP-AMP while
increasing the multiplicity of walking motions from $\times1$ to $\times100$.

As the walking multiplicity increases, AMP exhibits slower learning and lower
final performance, particularly under the $\times20$ and $\times100$
settings. In contrast, CMP-AMP maintains similar learning trajectories across
imbalance levels, supporting its reduced sensitivity to dominant motion
frequencies.

\section{Network Architectures}
\label{app:network_architectures}

\begin{figure}[t]
    \centering

    \begin{minipage}[c]{0.72\columnwidth}
        \centering
        \includegraphics[width=\linewidth]{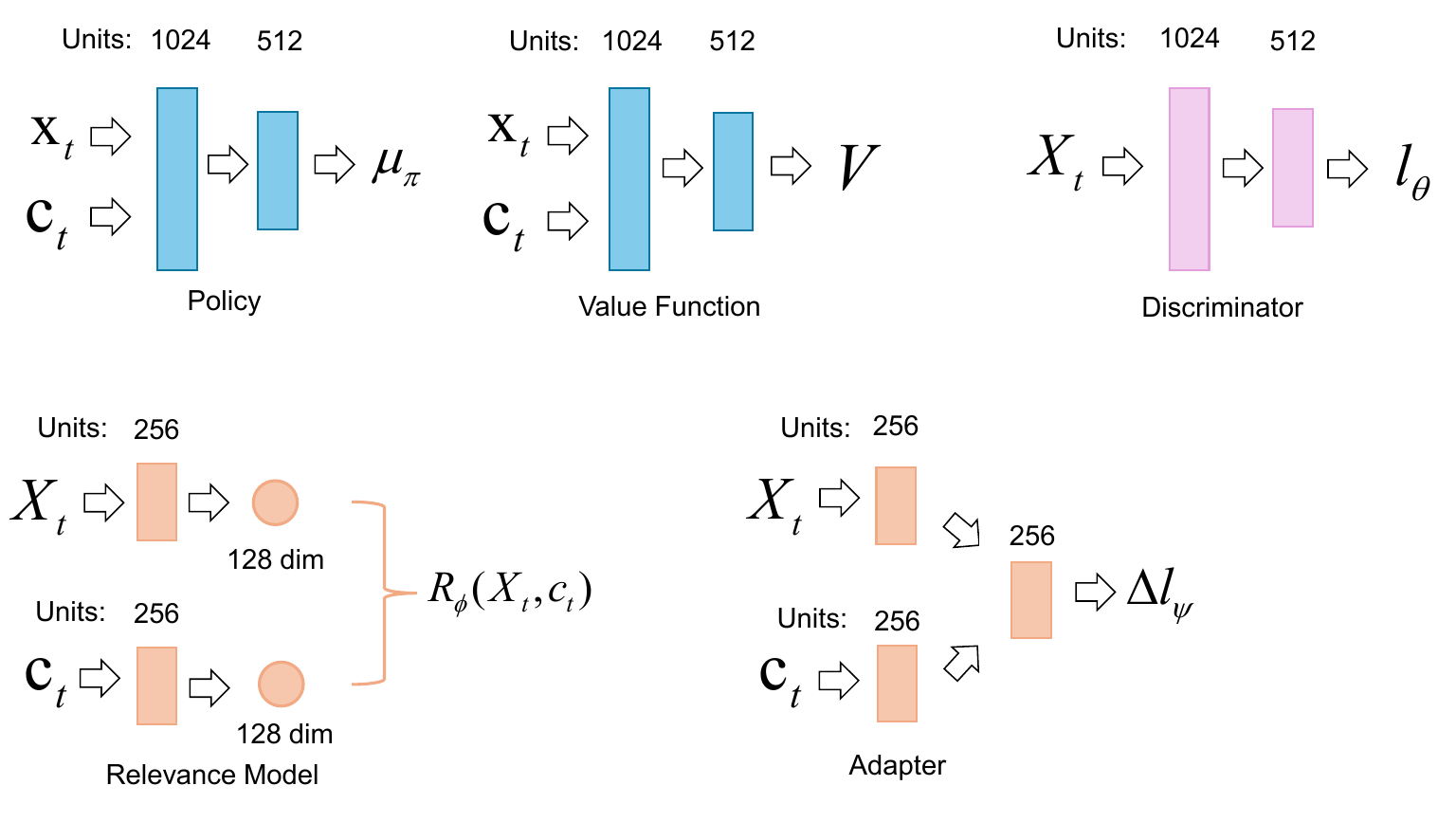}
    \end{minipage}
    \hfill
    \begin{minipage}[c]{0.26\columnwidth}
        \centering
        \includegraphics[width=\linewidth]{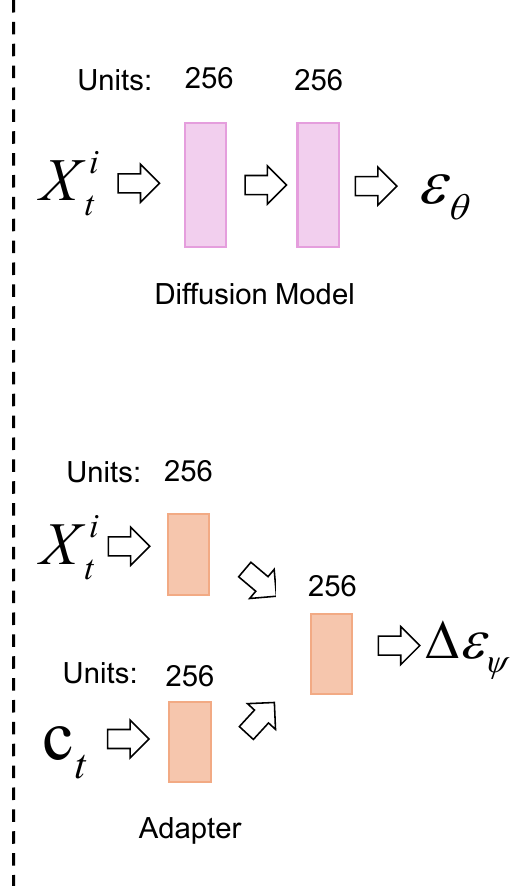}
    \end{minipage}

    \caption{\textbf{Network architectures used to model the components
    of CMP.}
    Left: the policy, value function, base AMP discriminator, contrastive
    relevance model, and residual logit adapter used in the primary
    CMP-AMP instantiation. These components are implemented using fully
    connected networks.
    Right: the CMP-SMP extension. The pretrained diffusion model is
    implemented using Transformer blocks with self-attention and adaptive
    normalization, while CMP adds a context-conditioned residual denoising
    adapter.}
    \label{fig:network_architectures}
\end{figure}

We use multilayer perceptrons for the policy, value function,
discriminator, relevance-model, and AMP-adapter components. Unless
otherwise specified, the same network architectures are used across all
tasks; only the input dimensions of the task-specific context, policy
observation, and value-function observation vary across environments.
Figure~\ref{fig:network_architectures} provides a schematic overview of
the architectures used in the AMP-based instantiation and the SMP-based
extension.

Let $s_t$ denote the task-agnostic character observation and $c_t$ the
task-specific context observation. The policy and value-function inputs
are formed from the corresponding environment observations, which combine
these two sources of information. The policy network contains two fully
connected hidden layers with $[1024,512]$ units, followed by a linear
output layer that predicts the action mean. The value function uses the
same hidden-layer architecture and a single linear output unit.

Following the motion representation used in MimicKit, let
$x_t=\Phi(s_t)$ denote the single-frame discriminator observation extracted
from environment state $s_t$. The mapping $\Phi$ retains motion-related
kinematic features while excluding task-specific context information. A
motion clip is formed from ten consecutive frames:
\[
X_t :=
\left(x_{t-8},\ldots,x_{t+1}\right).
\]
The clip $X_t$ is flattened and normalized before being provided to the
base AMP discriminator. The discriminator contains two fully connected
hidden layers with $[1024,512]$ units, followed by a single linear output
unit that produces the scalar motion logit.

The contrastive relevance model consists of a context encoder
$f_{\phi_c}$ and a motion encoder $g_{\phi_x}$. Each encoder contains one
fully connected hidden layer with 256 units and a SiLU activation,
followed by a linear projection to a 128-dimensional embedding. The
resulting embeddings are $\ell_2$-normalized, such that their dot product
corresponds to cosine similarity. The context encoder receives $c_t$,
while the motion encoder receives the same flattened and normalized motion
clip $X_t$ as the base AMP discriminator. The encoder architectures,
hidden width, and embedding dimension are shared across all tasks.

The CMP-AMP residual adapter contains separate motion and context branches.
Each branch uses a fully connected hidden layer with 256 units and a ReLU
activation, followed by a linear projection to a 256-dimensional feature.
The two features are concatenated and passed through a fusion network with
a 256-unit hidden layer and a scalar output,
$\Delta l_{\psi}(c_t,X_t)$. The final output layer is zero-initialized, so
the adapted discriminator initially reproduces the base AMP discriminator.

For the SMP-based extension, we retain the pretrained diffusion denoiser
architecture used by SMP. The denoiser first projects the noised motion
tokens into a 256-dimensional representation and processes them using four
Transformer blocks with multi-head self-attention, timestep-conditioned
adaptive normalization, and sinusoidal sequence positional embeddings.
The pretrained denoiser remains frozen during downstream training. CMP-SMP
adds a context-conditioned residual denoising adapter that receives the
noised motion clip, diffusion timestep, and task context. The adapter uses
sequence and timestep embeddings followed by a three-layer multilayer
perceptron with a hidden width of 512 and predicts a residual noise term
with the same dimensionality as the base denoiser output.

\section{Evaluation Metrics}
We report mean episodic task return and samples to threshold. For each of
three random seeds, task return is averaged over 32 evaluation episodes,
and the reported results are aggregated across seeds. The samples-to-threshold
metric is the number of environment interactions required for the mean
evaluation return to reach $80\%$ of the corresponding base prior's final
return. The same threshold is applied to the base method and its CMP variant
within each prior family. The two metrics measure task performance and
sample efficiency, respectively. Returns are compared only within each task,
and samples to threshold only within each prior family.
\begin{table}[t]
    \centering
    \caption{Return thresholds used to measure sample efficiency. Each
    threshold is set to $80\%$ of the corresponding base prior's final
    return and shared by both methods within that prior family.}
    \label{tab:return_thresholds}
    \begin{tabular}{lcc}
        \toprule
        \textbf{Task}
        & \textbf{AMP-based}
        & \textbf{SMP-based} \\
        \midrule
        Target Location      & $326$ & $300$ \\
        Steering             & $239$ & $396$ \\
        Trajectory Following & $147$ & $201$ \\
        Dodgeball            & $339$ & $357$ \\
        Dribbling            & $255$ & $366$ \\
        \bottomrule
    \end{tabular}
\end{table}

\section{Task Implementation}
\label{app:task_implementation}

\begin{figure*}[t]
    \centering
    \includegraphics[width=\textwidth]{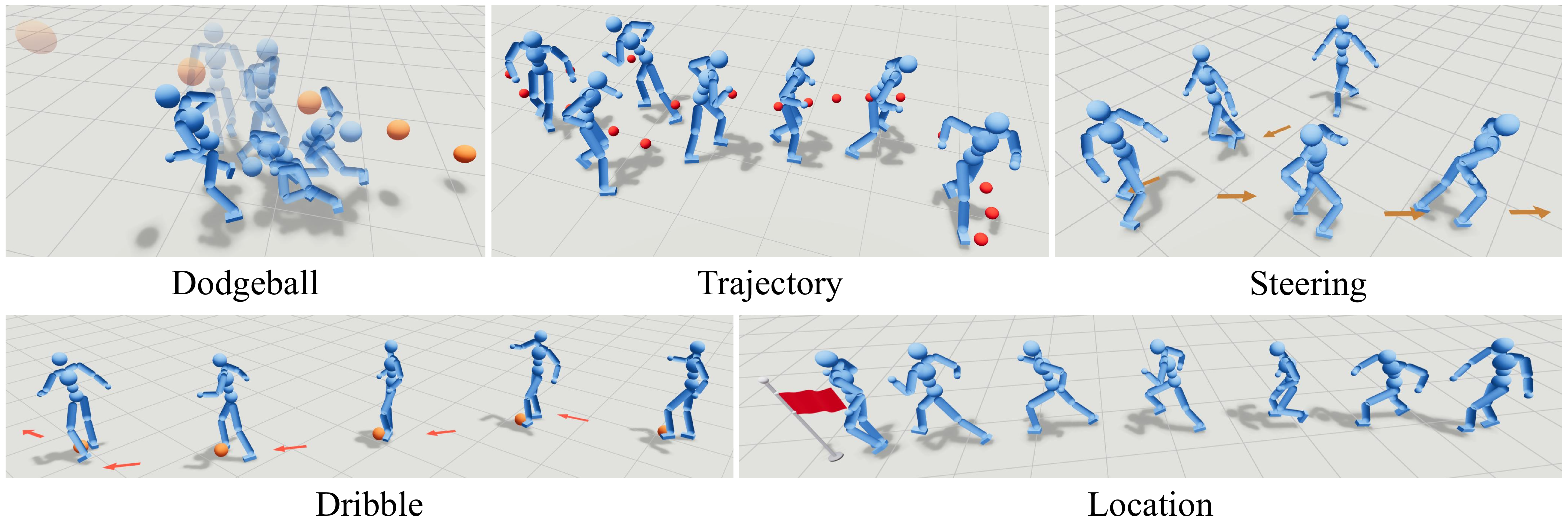}
    \caption{Visual overview of the five task implementations.}
    \label{fig:task_description}
\end{figure*}

We provide detailed descriptions of each task, its context definition,
and the task reward function used during training. We additionally
visualize the five task settings and their context definitions in
Figure~\ref{fig:task_description}.

\paragraph{Target Location.}
The character is required to move toward a target position
$\mathbf{x}^{*}_t$.

\emph{Context.}
The context is the target position expressed in the character's local
heading frame:
\[
c_t = \widetilde{\mathbf{x}}^{*}_t \in \mathbb{R}^{2}.
\]

\emph{Reward.}
The task reward encourages the character to approach the target, move
toward it at the desired speed, and face the target:
\[
\begin{aligned}
r_t^G
={}&
0.5
\exp\left(
-0.5
\left\|
\mathbf{x}^{*}_{t,xy}
-
\mathbf{x}^{\mathrm{root}}_{t,xy}
\right\|_2^2
\right)
\\
&+
0.4
\exp\left(
-4
\left[
\max\left(
0,
v^{*}
-
\mathbf{d}^{*}_t
\cdot
\dot{\mathbf{x}}^{\mathrm{root}}_{t,xy}
\right)
\right]^2
\right)
\\
&+
0.1
\max\left(
0,
\mathbf{d}^{*}_t
\cdot
\mathbf{f}_t
\right).
\end{aligned}
\]
Here, $\mathbf{d}^{*}_t$ is the horizontal unit vector pointing from the
character root to the target, $v^{*}$ is the desired target-directed
speed, and $\mathbf{f}_t$ is the character's horizontal facing direction.
The velocity reward is set to zero when the character moves away from the
target. Within the target threshold, the velocity and facing rewards are
set to one.

\paragraph{Steering.}
The character is required to follow a commanded horizontal velocity
while maintaining a desired facing direction.

\emph{Context.}
The context consists of the desired moving direction, target speed, and
desired facing direction, with both directions expressed in the
character's local heading frame:
\[
c_t =
\left[
\widetilde{\mathbf d}^{\mathrm{vel}}_t,\,
v_t^{*},\,
\widetilde{\mathbf d}^{\mathrm{face}}_t
\right]
\in \mathbb{R}^{5}.
\]

\emph{Reward.}
The task reward encourages the character to match the commanded velocity
and facing direction:
\[
\begin{aligned}
r_t^G
={}&
0.7
\exp\left(
-0.5
\left\|
v_t^{*}\mathbf d_t^{\mathrm{vel}}
-
\dot{\mathbf x}^{\mathrm{root}}_{t,xy}
\right\|_2^2
\right)
\\
&+
0.3
\max\left(
0,
{\mathbf d_t^{\mathrm{face}}}^{\top}\mathbf f_t
\right).
\end{aligned}
\]
Here, $\mathbf d_t^{\mathrm{vel}}$ and
$\mathbf d_t^{\mathrm{face}}$ denote the desired horizontal moving and
facing directions, respectively, and $\mathbf f_t$ is the character's
current horizontal facing direction. The velocity reward is set to zero
when the character moves opposite to the commanded direction.

\paragraph{Trajectory Following.}
At the beginning of each episode, a time-indexed planar trajectory
$\mathbf{x}^{*}(\tau)$ is generated and remains fixed throughout the
episode. The character is required to continuously track this trajectory
according to the current episode time.

\emph{Context.}
At every control step $t$, the context provides the current and future
trajectory points expressed relative to the character in its local
heading frame:
\[
c_t =
\left[
\widetilde{\mathbf{x}}^{*}(t),
\widetilde{\mathbf{x}}^{*}(t+\Delta),
\ldots,
\widetilde{\mathbf{x}}^{*}(t+(K-1)\Delta)
\right]
\in \mathbb{R}^{2K}.
\]
We use $K=6$ samples with $\Delta=0.30$ s. Since the first
sample corresponds to the current time, the furthest waypoint
lies $(K-1)\Delta=1.50$ s ahead. Although the trajectory remains
fixed within an episode, this context window advances at every control
step.

\emph{Reward.}
The task reward encourages the character to track the trajectory position,
match its local velocity, face along its tangent direction, and avoid
excessive speed:
\[
\begin{aligned}
r_t^G
={}&
0.5
\exp\left(
-7
\left\|
\mathbf{x}^{*}(t)_{xy}
-
\mathbf{x}^{\mathrm{root}}_{t,xy}
\right\|_2^2
\right)
\\
&+
0.3
\exp\left(
-5
\left\|
\mathbf{v}^{*}_t
-
\dot{\mathbf{x}}^{\mathrm{root}}_{t,xy}
\right\|_2^2
\right)
\\
&+
0.1
\max\left(
0,
{\mathbf{d}^{*}_t}^{\top}\mathbf{f}_t
\right)
\\
&+
0.1
\exp\left(
-3
\left[
\max\left(
0,
\left\|\dot{\mathbf{x}}^{\mathrm{root}}_{t,xy}\right\|_2
-
\left\|\mathbf{v}^{*}_t\right\|_2
-
0.3
\right)
\right]^2
\right).
\end{aligned}
\]
The target velocity and trajectory tangent are estimated using a
look-ahead interval $\Delta_r$:
\[
\mathbf{v}^{*}_t
=
\frac{
\mathbf{x}^{*}(t+\Delta_r)_{xy}
-
\mathbf{x}^{*}(t)_{xy}
}{
\Delta_r
},
\qquad
\mathbf{d}^{*}_t
=
\frac{\mathbf{v}^{*}_t}
{\left\|\mathbf{v}^{*}_t\right\|_2}.
\]
Here, $\mathbf{f}_t$ denotes the character's horizontal facing direction.
The velocity reward is set to zero when the character moves opposite to
the trajectory tangent.

\paragraph{Dodgeball.}
The character is required to dodge a ball repeatedly thrown toward its
torso. At each throw, the ball is initialized $8$--$10\,\mathrm{m}$ away
from the character and launched at a speed of $12$--$15\,\mathrm{m/s}$
toward the predicted torso position. The time interval between two
consecutive throws is randomly sampled from $1$ to $4\,\mathrm{s}$.

\emph{Context.}
The context consists of the ball's relative position and velocity,
expressed in the character's local heading frame:
\[
c_t =
\left[
\widetilde{\mathbf{x}}^{\mathrm{ball}}_t,\,
\widetilde{\mathbf{v}}^{\mathrm{ball}}_t
\right]
\in \mathbb{R}^{6}.
\]

\emph{Reward.}
The task reward encourages the character to keep away from the ball while
avoiding unnecessary horizontal motion:
\[
\begin{aligned}
r_t^G
={}&
0.9
\left[
1-
\exp\left(
-0.3
\left\|
\mathbf{x}^{\mathrm{ball}}_t
-
\mathbf{x}^{\mathrm{root}}_t
\right\|_2
\right)
\right]
\\
&+
0.1
\exp\left(
-
\left\|
\dot{\mathbf{x}}^{\mathrm{root}}_{t,xy}
\right\|_2^2
\right).
\end{aligned}
\]
The episode terminates when the ball hits the character, as detected by
physical contact or a sudden change in the ball's velocity near the
character.
\paragraph{Dribbling.}
The character is required to keep the ball close to its feet while moving
it along a sampled horizontal direction at a desired speed.

\emph{Context.}
The context consists of the ball's relative position and velocity, the
desired moving direction, and the target speed, all directional quantities
being expressed in the character's local heading frame:
\[
c_t =
\left[
\widetilde{\mathbf{x}}^{\mathrm{ball}}_t,\,
\widetilde{\mathbf{v}}^{\mathrm{ball}}_t,\,
\widetilde{\mathbf{d}}^{*}_t,\,
v^{*}_t
\right]
\in \mathbb{R}^{9}.
\]

\emph{Reward.}
The task reward encourages the character to remain close to the ball,
match the commanded ball velocity, and keep the ball near the ground:
\[
\begin{aligned}
r_t^G
={}&
0.35
\exp\left(
-4
\min_{i\in\mathcal{F}}
\left\|
\mathbf{x}^{\mathrm{foot},i}_{t,xy}
-
\mathbf{x}^{\mathrm{ball}}_{t,xy}
\right\|_2^2
\right)
\\
&+
0.55
\exp\left(
-2.5
\left\|
v_t^{*}\mathbf{d}^{*}_t
-
\dot{\mathbf{x}}^{\mathrm{ball}}_{t,xy}
\right\|_2^2
\right)
\\
&+
0.10
\exp\left(
-8
\left(
x^{\mathrm{ball}}_{t,z}
-
h_{\mathrm{ball}}
\right)^2
\right),
\end{aligned}
\]
where $\mathcal{F}$ contains the left and right feet,
$\mathbf{d}^{*}_t$ is the desired horizontal ball direction,
$v_t^{*}$ is the target speed, and $h_{\mathrm{ball}}$ is the nominal
ball height. The velocity reward is set to zero when the ball moves
opposite to the desired direction.

The episode terminates when the horizontal character--ball distance
exceeds $6\,\mathrm{m}$ or the ball height exceeds $1.2\,\mathrm{m}$,
after an initial grace period of $0.5\,\mathrm{s}$.

\section{Training Hyperparameters}
\label{app:training_hyperparameters}

We report the hyperparameters for the AMP-based and SMP-based
instantiations separately. Table~\ref{tab:amp_hyperparameters}
summarizes the optimization and motion-prior settings shared by AMP
and CMP-AMP, while Table~\ref{tab:cmp_amp_hyperparameters} reports
the additional residual-adapter and contrastive-relevance settings
used by CMP-AMP. Analogously,
Table~\ref{tab:smp_hyperparameters} reports the settings shared by
SMP and CMP-SMP, and Table~\ref{tab:cmp_smp_hyperparameters}
summarizes the additional settings introduced by CMP-SMP. Unless
otherwise specified, the same settings are used across all tasks
without task-specific tuning.

Following the MimicKit implementation, the configured actor, critic,
discriminator, and adapter batch-size parameters are per-environment
multipliers and are scaled by the number of parallel environments
$N_{\mathrm{env}}$. In contrast, the contrastive, demonstration, and
SMP evaluation batch sizes are absolute and independent of
$N_{\mathrm{env}}$.

\begin{table}[t]
\centering
\caption{Hyperparameters shared by AMP and CMP-AMP.}
\label{tab:amp_hyperparameters}
\small
\setlength{\tabcolsep}{4pt}
\renewcommand{\arraystretch}{0.96}
\begin{tabular}{lc}
\toprule
Parameter & Value \\
\midrule

\multicolumn{2}{l}{\textit{Policy optimization}} \\
Task/prior weights; rollout steps
& $0.5/0.5;\ 32$ \\
Policy/value learning rates
& $2{\times}10^{-5}/5{\times}10^{-5}$ \\
Policy/value epochs (batch mult.)
& $5/2\;(4/2)$ \\
Action std. / $\gamma$ / GAE $\lambda$
& $0.05/0.99/0.95$ \\
PPO / advantage clip / bound weight
& $0.2/4.0/10.0$ \\

\midrule
\multicolumn{2}{l}{\textit{AMP discriminator}} \\
Learning rate / weight decay
& $2.5{\times}10^{-4}/10^{-4}$ \\
Epochs / replay size / samples
& $2/(2{\times}10^{5})/1000$ \\
Logit reg. / gradient penalty / reward scale
& $0.01/5.0/2.0$ \\

\bottomrule
\end{tabular}
\end{table}

\begin{table}[!htbp]
\centering
\caption{CMP-AMP-specific training hyperparameters.}
\label{tab:cmp_amp_hyperparameters}
\small
\setlength{\tabcolsep}{5pt}
\renewcommand{\arraystretch}{1.06}
\begin{tabular}{lc}
\toprule
Parameter & Value \\
\midrule

\multicolumn{2}{l}{\textit{Residual adapter and schedule}} \\
Adapter learning rate & $5\times10^{-5}$ \\
Residual scale $\lambda_{\mathrm{res}}$ & $0.03$ \\
Residual regularization $\eta$ & $0.01$ \\
Adapter-loss weight & $0.5$ \\

\midrule
\multicolumn{2}{l}{\textit{Contrastive relevance learning}} \\
Learning rate & $10^{-4}$ \\
Contrastive-loss weight & $0.1$ \\
Temperature $\tau$ & $0.1$ \\
Contrastive and demonstration batch size & $512$ \\

\midrule
\multicolumn{2}{l}{\textit{Positive construction}} \\
Minimum positives / fallback fraction & $64/0.35$ \\
Advantage temperature $\beta_{\mathrm{adv}}$ & $1.0$ \\
Demonstration weight $\lambda_{\mathrm{demo}}$ & $0.5$ \\

\midrule
\multicolumn{2}{l}{\textit{Reference weighting}} \\
Reweighting scale $\alpha$ & $0.5$ \\
Weight range $[w_{\min},w_{\max}]$ & $[0.5,2.0]$ \\

\bottomrule
\end{tabular}
\end{table}

\begin{table}[!htbp]
\centering
\caption{Hyperparameters shared by SMP and CMP-SMP.}
\label{tab:smp_hyperparameters}
\small
\setlength{\tabcolsep}{4pt}
\renewcommand{\arraystretch}{1.0}
\begin{tabular}{lc}
\toprule
Parameter & Value \\
\midrule

\multicolumn{2}{l}{\textit{Policy optimization}} \\
Task/prior weights; rollout steps
& $0.5/0.5;\ 32$ \\
Policy/value learning rates
& $10^{-4}/10^{-4}$ \\
Policy/value epochs (batch mult.)
& $5/2\;(4/2)$ \\
Action std. / $\gamma$ / GAE $\lambda$
& $0.05/0.99/0.95$ \\
PPO / advantage clip / bound weight
& $0.2/4.0/10.0$ \\

\midrule
\multicolumn{2}{l}{\textit{Score-matching motion prior}} \\
Diffusion steps $\mathcal{K}$ / eval. batch
& $\{22,15,8\}/4096$ \\
Reward scale $\kappa$ / normalizer samples
& $6.0/10^{8}$ \\

\bottomrule
\end{tabular}
\end{table}

\begin{table}[t]
\centering
\caption{CMP-SMP-specific training hyperparameters.}
\label{tab:cmp_smp_hyperparameters}
\small
\setlength{\tabcolsep}{5pt}
\renewcommand{\arraystretch}{1.06}
\begin{tabular}{lc}
\toprule
Parameter & Value \\
\midrule

\multicolumn{2}{l}{\textit{Residual adapter and schedule}} \\
Adapter learning rate & $10^{-4}$ \\
Residual scale $\lambda_{\mathrm{res}}$ & $0.1$ \\
Residual regularization $\eta$ & $0.001$ \\
Contrastive / adapter warmup iterations & $5/10$ \\

\midrule
\multicolumn{2}{l}{\textit{Contrastive relevance learning}} \\
Learning rate & $10^{-4}$ \\
Contrastive-loss weight & $0.1$ \\
Temperature $\tau$ & $0.1$ \\
Contrastive and demonstration batch size & $512$ \\

\midrule
\multicolumn{2}{l}{\textit{Positive construction}} \\
Minimum positives / fallback fraction & $64/0.35$ \\
Advantage temperature $\beta_{\mathrm{adv}}$ & $1.0$ \\
Demonstration weight $\lambda_{\mathrm{demo}}$ & $1.0$ \\

\midrule
\multicolumn{2}{l}{\textit{Reference weighting}} \\
Reweighting scale $\alpha$ & $1.0$ \\
Weight range $[w_{\min},w_{\max}]$ & $[0.5,2.0]$ \\

\bottomrule
\end{tabular}
\end{table}

\section{Experimental Details}
\label{app:experimental_details}

Isaac Gym is the simulator; PhysX is the physics engine. We report the hardware and software configurations used for
training and evaluation in Table~\ref{tab:experimental_details}.

\begin{table}[t]

\centering
\caption{Experimental environment details.}
\begin{tabular}{ll}

\toprule
Component & Configuration \\
\midrule
Simulator & NVIDIA Isaac Gym Preview 4 \\
Physics Engine & PhysX \\
GPU & NVIDIA RTX 3090 / RTX 4090 \\
CPU & AMD EPYC 7543 \\
System memory & 512 GB \\
Operating system & Ubuntu 20.04.1 LTS \\
CUDA & 12.1 \\
Python & 3.8 \\
Precision & FP32 \\
Parallel Environments & 4096 \\
Evaluation Seeds & 3 \\
\bottomrule
\label{tab:experimental_details}
\end{tabular}
\end{table}


\end{document}